%% file: main.tex
\documentclass{bmvc2k}
\usepackage{cleveref}
\usepackage{amsfonts}
\usepackage{amsthm}
\usepackage{booktabs}
\usepackage{multirow}
\usepackage{graphicx}
\usepackage{pifont}
\newcommand{\cmark}{\ding{51}}
\usepackage[table]{xcolor}
\usepackage{algorithm}
\usepackage{algpseudocode}
\newtheorem{theorem}{Theorem}
\usepackage{xcolor}
\definecolor{attackblue}{RGB}{20,80,160}
\newcommand{\att}[1]{\textcolor{attackblue}{\textbf{#1}}}

\title{Ranking vs. Assignment: The Metric Mismatch in Multi-View Object Association}

\addauthor{Matvei Shelukhan}{matveishelukhan@gmail.com}{1*}
\addauthor{Timur Mamedov}{me@timmzak.com}{1, 2*}
\addauthor{Aleksandr Chukhrov}{aleksandr.chukhrov@tevian.ru}{1}
\addauthor{Karina Kvanchiani}{karinakvanciani@gmail.com}{1}

\addinstitution{
 Tevian\\
 Moscow, Russia
}
\addinstitution{
 Lomonosov Moscow State University\\
 Moscow, Russia
}

\runninghead{Shelukhan, Mamedov, Chukhrov, Kvanchiani}{Ranking vs. Assignment}

\begin{document}

\maketitle
\input{sec/0_abstract}
\input{sec/1_intro}
\input{sec/2_rel_work}
\input{sec/3_method}
\input{sec/4_exps}
\input{sec/5_conclusion}
\bibliography{egbib}

\newpage
\appendix
\input{sec/6_supplementary}

\end{document}

%% file: sec/0_abstract.tex
\begin{abstract}
Multi-view object association is an important computer vision problem that underlies many multi-camera perception tasks. While this task is naturally formulated as a constrained one-to-one matching problem, recent works heavily rely on pairwise ranking metrics like AP and FPR-95 for model evaluation. We highlight a fundamental mismatch between these metrics and the actual assignment objective. Theoretically, we show that AP and FPR-95 can be imperfect even when the assignment is already correct, and that Sinkhorn-based normalization can make them perfect. Conversely, optimal pairwise ranking can still lead to incorrect assignments. We validate this mismatch in practice by using our Sinkhorn-based normalization as a controlled post-processing stress test. We show that optimizing just a few post-processing parameters significantly boosts AP and FPR-95 without corresponding improvements in assignment-level metrics such as ACC and IPAA.
\end{abstract}

%% file: sec/1_intro.tex
\section{Introduction}
\label{sec:intro}
Multi-view object association aims to match detections of the same physical object across synchronized camera views. Given a set of detections in each view, the goal is to establish one-to-one cross-view correspondences. Each detection can be associated with at most one detection in another view, while objects may remain unmatched in views where they are not visible. Reliable multi-view association is essential for multi-camera perception systems, where a consistent understanding of the scene requires merging observations from multiple viewpoints. It appears in applications such as multi-camera tracking~\cite{nguyen2022lmgp}, robotic perception~\cite{bandi2025action}, sports analytics~\cite{liang2020multi}, and human activity understanding~\cite{al2025structured}.

Despite the importance of the task, evaluation protocols for multi-view object association remain heterogeneous. Existing works~\cite{chen2025learning, cai2020messytable, park2023mvdet, seo2023vit} report a mixture of metrics, including pairwise ranking metrics such as Average Precision (AP) and False Positive Rate at 95\% True Positive Rate (FPR-95), and assignment-level metrics such as Accuracy (ACC) and Image-Pair Association Accuracy (IPAA). AP and FPR-95 are natural choices for evaluating pairwise affinity matrices, as they measure whether true cross-view pairs receive higher scores than false pairs. Accordingly, these metrics have been adopted in recent multi-view association methods.

\begin{figure*}[t]
  \centering
  \includegraphics[width=0.5\linewidth]{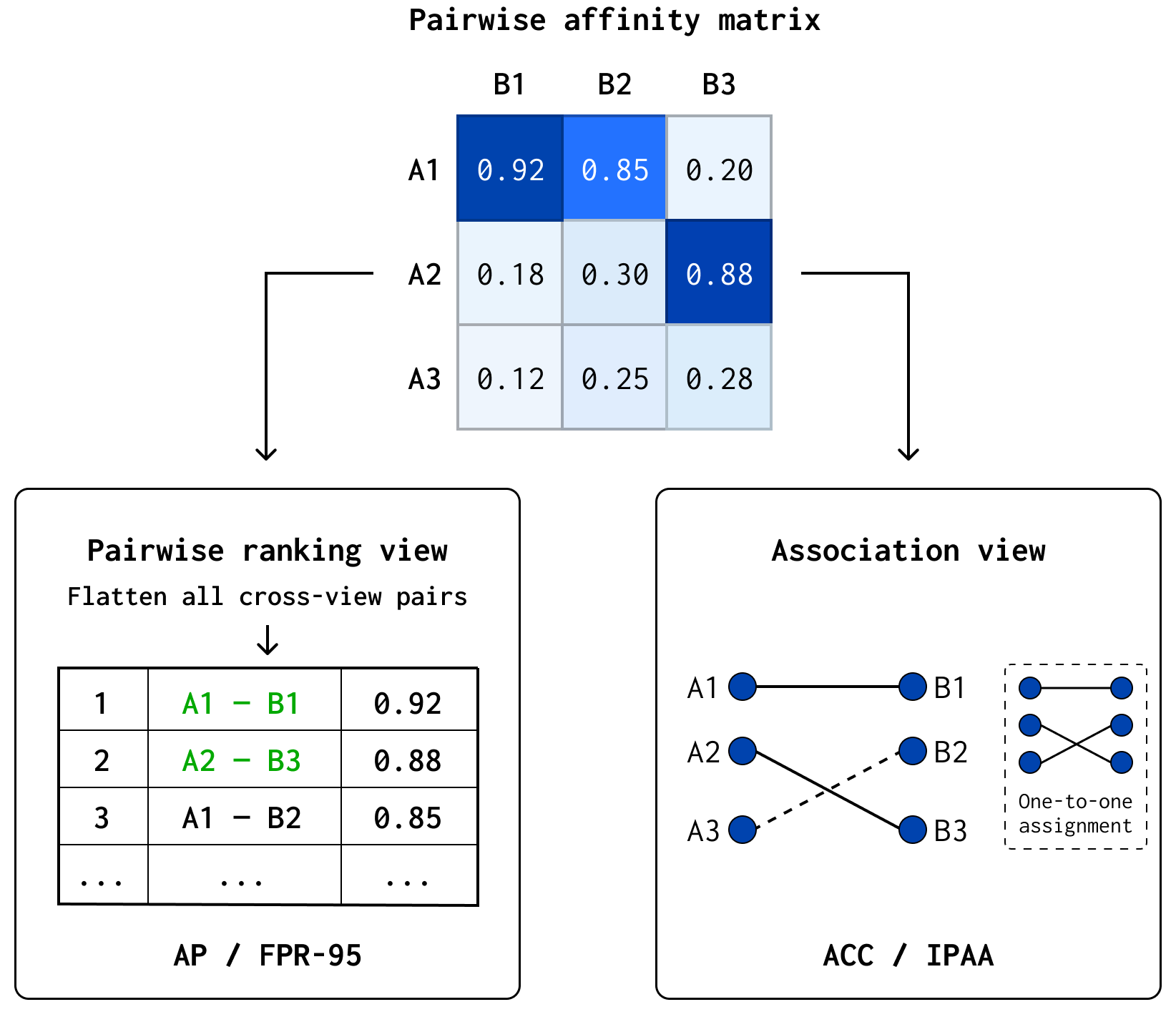}
  \caption{
    Pairwise ranking metrics and assignment-level metrics evaluate different aspects of the affinity matrix.
    AP and FPR-95 flatten all cross-view pairs and measure their ranking quality, while ACC and IPAA evaluate the one-to-one matching obtained from these scores.
    }
  \label{fig:mismatch}
\end{figure*}
However, multi-view association is not merely a pairwise ranking problem. Since the final output is a constrained one-to-one matching, assignment-level metrics are needed to evaluate the actual association decisions. Pairwise ranking metrics therefore create a fundamental mismatch when used as primary indicators of association quality (\Cref{fig:mismatch}).

In this paper, we show that this mismatch is not only conceptual but also exploitable. We use Sinkhorn-based normalization~\cite{cuturi2013sinkhorn} as a controlled post-processing stress test to examine whether evaluation metrics are sensitive to score reshaping. We show that learning only a small number of post-processing parameters is sufficient to substantially improve AP and FPR-95 while leaving the underlying model unchanged. Crucially, these improvements in pairwise ranking metrics have only a limited effect on assignment-level metrics such as ACC and IPAA. Our theoretical analysis shows that this issue appears in both directions. First, AP and FPR-95 can be imperfect even when the assignment is correct. In this case, Sinkhorn-based normalization can improve these pairwise metrics to perfect values. Second, optimal pairwise ranking can still coexist with a completely wrong assignment.


\vspace{2pt}\noindent This paper makes the following contributions:
\begin{itemize}
    \item We analyze the mismatch between pairwise ranking metrics and constrained one-to-one matching in multi-view object association.
    \item Theoretically, we show that AP and FPR-95 can be imperfect even when the assignment is correct, and that Sinkhorn-based normalization can make them perfect. Conversely, optimal pairwise ranking can still lead to wrong assignments.
    \item We evaluate the robustness of common metrics under our Sinkhorn-based post-processing stress test. Using multi-view person association as a representative setting, we show across multiple association methods that AP and FPR-95 can be substantially improved by tuning only post-processing parameters, without corresponding gains in Accuracy and IPAA.
\end{itemize}
The main empirical study focuses on multi-view person association, where a broad set of public methods and evaluation protocols is available. Since our analysis operates directly on pairwise affinity matrices, it is agnostic to the object category. We additionally validate this on the MessyTable~\cite{cai2020messytable} object association benchmark, where the same mismatch is observed (see supplementary material).

%% file: sec/2_rel_work.tex
\section{Related Work}
\label{sec:rel_work}

\subsection{Multi-view Object and Person Association}
Multi-view object association aims to identify object instances that correspond to the same physical object across synchronized camera views. MessyTable~\cite{cai2020messytable} studied object association in cluttered multi-view tabletop scenes and showed that reliable association requires combining appearance, context, and geometric cues. ViT-P3DE~\cite{seo2023vit} further addressed multi-camera object association with a vision transformer architecture and pseudo 3D position embeddings.

A more extensively studied special case is multi-view person association. Person re-identification models~\cite{zhou2021learning, mamedov2025dynamix} provide strong appearance descriptors, but appearance alone can be unreliable in crowded scenes. GNN-CCA~\cite{luna2022graph} and ReST~\cite{cheng2023rest} addressed multi-camera association with graph-based models. MvMHAT~\cite{gan2021self} considered the joint setting of multi-view multi-human association and tracking, using self-supervised learning to exploit spatial and temporal consistency. More recently, Self-MVA~\cite{chen2025learning} proposed a self-supervised uncalibrated multi-view person association framework that learns geometric features without requiring manual identity annotations.

Related works have also studied robustness to imperfect cross-view correspondences. Multi-view clustering methods address partially unaligned, missing, or noisy cross-view data~\cite{yang2022robust,zhou2026uncover}, while DART~\cite{yang2022learning} considers noisy annotations and cross-modal correspondences in visible-infrared person re-identification.

The multi-view association methods discussed above differ in supervision, architecture, and use of geometry, but they share a common interface. They produce pairwise affinity matrices that are converted into one-to-one cross-view matches. This makes multi-view association a natural setting for studying how pairwise affinity quality relates to the quality of the induced assignment.

\subsection{Evaluation Metrics}
Evaluation protocols for multi-view object association combine pairwise ranking metrics and assignment-level metrics. Pairwise metrics such as Average Precision (AP) and False Positive Rate at 95\% True Positive Rate (FPR-95) evaluate the affinity matrix before enforcing a matching constraint. These metrics are natural for evaluating ranking or retrieval quality and are commonly reported in multi-view association works~\cite{cai2020messytable, seo2023vit, park2023mvdet, chen2025learning}.

Since the final output of multi-view association is not a ranked list of pairs but a constrained matching, evaluation protocols also include metrics that assess the final predicted matches. MessyTable~\cite{cai2020messytable} introduced Image-Pair Association Accuracy (IPAA) to evaluate association quality at the level of an image pair. Self-MVA~\cite{chen2025learning} further complemented AP, FPR-95, and IPAA with Precision, Recall, and Accuracy (ACC) computed from the predicted cross-view matches. These assignment-level metrics evaluate the correctness and coverage of the final matching, while ACC additionally accounts for correctly unmatched identities.

This coexistence of metric families reflects a mismatch between pairwise affinity ranking and constrained association quality. Although Self-MVA~\cite{chen2025learning} noted that AP and FPR-95 do not fully capture association capability, the robustness of these metrics to assignment-aware score transformations has not been systematically studied. Our work addresses this gap by studying whether pairwise ranking metrics can be manipulated through post-processing without corresponding improvements in assignment-level quality.

\subsection{Sinkhorn Normalization}
One-to-one matching problems are commonly solved by converting pairwise affinities into a constrained assignment. The Hungarian algorithm~\cite{munkres1957algorithms} provides a classical discrete solution, while Sinkhorn normalization offers a differentiable relaxation by transforming an affinity matrix into a doubly-stochastic form~\cite{cuturi2013sinkhorn}. Sinkhorn normalization has also been widely adopted in vision and retrieval models. SuperGlue~\cite{sarlin2020superglue} used a Sinkhorn-based optimal matching layer for local feature matching. In retrieval, Sinkhorn-style normalization has been used to balance assignment probabilities~\cite{pan2025hubness}. In representation learning and person re-identification, optimal-transport-based transformations have been used to refine features~\cite{shalam2022self} and guide cross-modality matching or pseudo-label assignment~\cite{wang2022optimal,zhang2024mutual}.

Recent works have addressed partial and unreliable correspondences through optimal transport and related matching formulations. Norton~\cite{lin2024multi} combines Sinkhorn normalization with an alignable prompt bucket to filter out unalignable video-text elements. Other approaches handle partial correspondences through learnable outlier rejection in graph matching~\cite{lin2026learning}, masked optimal transport for partial domain adaptation~\cite{luo2023mot}, or optimal partial transport that allows only part of the mass to be transported~\cite{bai2023sliced}.

Our use of Sinkhorn normalization differs from these works. Prior methods typically employ Sinkhorn, optimal transport, or related matching mechanisms as part of the model or training objective to improve matching, retrieval, or representation quality. In contrast, we use Sinkhorn normalization as part of our controlled post-processing stress test on an already computed affinity matrix.

%% file: sec/3_method.tex
\section{Preliminaries}
\label{sec:preliminaries}

\subsection{Problem Setup and Evaluation Metrics}
\label{sec:metrics}
We consider synchronized images captured from multiple camera views. For timestamp \(t\) and view \(c\), let
\begin{equation}
    \mathcal{D}_{t,c}=\{d_{t,c}^{1},\ldots,d_{t,c}^{n_{t,c}}\}
\end{equation}
be the detected objects, with identity labels \(y_{t,c}^{i}\in\mathcal{Y}_{t,c}\). For a view pair \((a,b)\) and timestamp \(t\), a multi-view association model outputs an affinity matrix
\begin{equation}
    S^{t,a,b}\in\mathbb{R}^{n_{t,a}\times n_{t,b}},
\end{equation}
where larger \(S^{t,a,b}_{ij}\) indicates that detections \(d_{t,a}^{i}\) and \(d_{t,b}^{j}\) are more likely to correspond to the same object. For readability, we omit timestamps and view indices and denote an affinity matrix by \(S\). Each \(S_{ij}\) is labeled positive when the corresponding detections share the same identity and negative otherwise.

Let \(\{(s_q,z_q)\}_{q=1}^{N}\) be all flattened scores and binary labels from all evaluated image pairs, sorted by decreasing score as \(s_{\pi(1)}\geq\cdots\geq s_{\pi(N)}\). 
Average Precision evaluates whether positive pairs are ranked above negative pairs:
\begin{equation}
    \mathrm{AP}
    =
    \frac{1}{N_+}
    \sum_{k=1}^{N}
    \left(
    \frac{1}{k}\sum_{\ell=1}^{k} z_{\pi(\ell)}
    \right) \cdot
    z_{\pi(k)},
    \qquad
    N_+=\sum_{q=1}^{N} z_q .
\end{equation}
False Positive Rate at 95\% True Positive Rate is defined as
\begin{equation}
    \mathrm{FPR}\text{-}95
    =
    \min_{\gamma:\,\mathrm{TPR}(\gamma)\geq0.95}
    \mathrm{FPR}(\gamma),
\end{equation}
where
\begin{equation}
    \mathrm{TPR}(\gamma)=
    \frac{\sum_q \mathbf{1}[s_q\geq\gamma]\cdot z_q}{\sum_q z_q},
    \qquad
    \mathrm{FPR}(\gamma)=
    \frac{\sum_q \mathbf{1}[s_q\geq\gamma] \cdot (1-z_q)}{\sum_q(1-z_q)}.
\end{equation}
Both AP and FPR-95 operate on flattened pairwise scores before enforcing any one-to-one matching constraint.

Assignment-level metrics first convert the affinity matrix into a predicted matching. 
Following prior evaluation protocols, we compute a maximum-weight bipartite matching with the Hungarian algorithm and then discard matches below a threshold \(\theta\):
\begin{equation}
    \mathcal{M}(S)
    =
    \{(i,j)\in \mathrm{Hungarian}(S): S_{ij}\geq \theta\}.
\end{equation}
A predicted match is correct if \(y_i^1=y_j^2\). For an image pair, let \(C\) be the number of correct predicted matches, \(P_{\mathrm{tot}}=|\mathcal{M}|\), and
\begin{equation}
    R_{\mathrm{tot}} = |\mathcal{I}_1\cap\mathcal{I}_2|,
    \qquad
    A_{\mathrm{tot}} = |\mathcal{I}_1\cup\mathcal{I}_2|,
\end{equation}
where \(\mathcal{I}_1\) and \(\mathcal{I}_2\) are the identity sets in the two views. Precision and Recall are aggregated over all image pairs as
\begin{equation}
    \mathrm{Precision}=\frac{\sum C}{\sum P_{\mathrm{tot}}},
    \qquad
    \mathrm{Recall}=\frac{\sum C}{\sum R_{\mathrm{tot}}}.
\end{equation}

Accuracy and IPAA also account for identities that correctly remain unmatched. 
Let \(T\) be the number of correctly unmatched identities, i.e., identities present in only one view and not selected in the predicted matching. 
Then the association accuracy is
\begin{equation}
    \mathrm{ACC}
    =
    \frac{\sum (C+T)}{\sum A_{\mathrm{tot}}}.
\end{equation}
For each image pair \(p\), define
\begin{equation}
    r_p=\frac{C^p+T^p}{A_{\mathrm{tot}}^p}.
\end{equation}
Image-Pair Association Accuracy at threshold \(X\) is
\begin{equation}
    \mathrm{IPAA}\text{-}X
    =
    \frac{1}{|\mathcal{P}|}
    \sum_{p\in\mathcal{P}}
    \mathbf{1}
    \left[
        r_p\geq \frac{X}{100}
    \right].
\end{equation}
Thus, ACC aggregates correct association decisions over all image pairs, while IPAA-\(X\) counts the fraction of image pairs whose association accuracy exceeds a predefined threshold, commonly \(80\%\), \(90\%\), or \(100\%\).

\subsection{Sinkhorn Normalization}
\label{sec:sinkhorn_prelim}
Sinkhorn normalization transforms an affinity matrix into a soft assignment matrix.
For an affinity matrix \(S\in\mathbb{R}^{n\times m}\) and temperature \(\tau>0\), define the positive kernel
\begin{equation}
    G_{\tau}(S)_{ij} = \exp(S_{ij}/\tau).
\end{equation}
The Sinkhorn-normalized matrix is obtained by alternately normalizing the rows and columns of \(G_{\tau}(S)\) for \(K\) iterations:
\begin{equation}
    P_{\tau}
    =
    \mathrm{Sinkhorn}_{K}\!\left(G_{\tau}(S)\right).
\end{equation}
Lower values of \(\tau\) produce sharper matrices, while larger values produce smoother matrices. After normalization, \(P_{\tau}\) is approximately doubly stochastic, so rows and columns compete for assignment mass. Thus, Sinkhorn normalization can be viewed as a soft relaxation of one-to-one bipartite matching. Notably, this transformation leaves the detections, features, and model parameters fixed. It only reshapes the affinity matrix before pairwise ranking or assignment-level metrics are computed.

\section{Theoretical Analysis}
\label{sec:theory}

We now formalize two complementary limitations of pairwise ranking metrics in multi-view association. First, AP and FPR-95 can be imperfect even when the Hungarian matching already recovers the correct assignments. In this case, Sinkhorn-based normalization can make these pairwise metrics perfect. The second result shows the converse limitation: even optimal pairwise ranking does not guarantee correct assignment quality. Together, these results demonstrate that AP and FPR-95 are not sufficient indicators of association performance in a constrained matching task. Detailed proofs of both theorems are provided in the supplementary material.

\subsection{Correct Assignment Does Not Imply Perfect Pairwise Metrics}
We consider the two-view association case with \(m\) detections in one view, \(n\) detections in the other view, and \(k\) identities shared across both views. Let \(S\in\mathbb{R}^{m\times n}\) be an affinity matrix and let \(M^\star\subset[m]\times[n]\) be the ground-truth matching of size \(k\). 
Assume that \(M^\star\) is the unique maximum-weight \(k\)-matching:
\begin{equation}
    \sum_{(i,j)\in M^\star} S_{ij}
    >
    \sum_{(i,j)\in M} S_{ij}
    \qquad
    \forall M\in\mathcal{M}_k,\; M\neq M^\star .
\end{equation}
Thus, the constrained matching problem already recovers the correct assignment.

\begin{theorem}
\label{thm:sinkhorn_perfect}
Under the condition above, there exists a Sinkhorn-based transformation \(T_\tau\), such that
\begin{equation}
    T_\tau(S) \rightarrow X^\star
    \quad
    \text{as}
    \quad
    \tau\rightarrow0^+,
\end{equation}
where \(X^\star_{ij}=\mathbf{1}[(i,j)\in M^\star]\). 
Consequently, for all sufficiently small \(\tau>0\),
\begin{equation}
    \mathrm{AP}(T_\tau(S))=1,
    \qquad
    \mathrm{FPR}\text{-}95(T_\tau(S))=0.
\end{equation}
\end{theorem}
The condition above does not imply that the original pairwise ranking is perfect. There exist matrices satisfying it in which a negative pair is ranked above a positive pair, so AP and FPR-95 are imperfect before normalization. Thus, the theorem shows that Sinkhorn-based normalization can make pairwise metrics perfect in a setting where the assignment is already correct, without degrading association quality.

In practice, however, association models produce imperfect affinity matrices that do not satisfy this idealized condition. As a result, ACC and IPAA are generally below one, and the effect of Sinkhorn normalization on assignment-level metrics becomes harder to predict. This motivates our post-processing stress test~(\Cref{sec:attack}), which empirically tests whether AP and FPR-95 can be improved without corresponding gains in assignment-level quality.

\subsection{Perfect Pairwise Ranking Does Not Imply Correct Assignment}
The second result shows that the mismatch can occur in the opposite direction. 
Even if pairwise ranking is perfect, the assignment induced by Hungarian matching can be completely wrong.

Consider two views containing \(N\) shared identities \(A_0,\ldots,A_{N-1}\). 
In addition, the first view contains one extra identity \(X\), and the second view contains one extra identity \(U\). 
The extra identities are not shared across views and therefore should remain unmatched. 
Thus, the only ground-truth cross-view matches are
\begin{equation}
    (A_i,A_i), \qquad i=0,\ldots,N-1.
\end{equation}
We now construct an affinity matrix \(S\in\mathbb{R}^{(N+1)\times(N+1)}\). 
Its rows correspond to \((A_0,\ldots,A_{N-1},X)\) in the first view, and its columns correspond to \((A_0,\ldots,A_{N-1},U)\) in the second view.

\begin{theorem}
\label{thm:perfect_ap_wrong_assignment}
For any \(N\geq 1\), there exists an affinity matrix
\(S\in\mathbb{R}^{(N+1)\times(N+1)}\) such that every true pair has a higher score than every false pair. 
Therefore,
\begin{equation}
    \mathrm{AP}(S)=1,
    \qquad
    \mathrm{FPR}\text{-}95(S)=0.
\end{equation}
Nevertheless, Hungarian matching selects only false matches. 
For a valid range of thresholds, the resulting assignment-level metrics satisfy
\begin{equation}
    \mathrm{Precision}=0,
    \qquad
    \mathrm{Recall}=0,
    \qquad
    \mathrm{ACC}=0,
    \qquad
    \mathrm{IPAA}\text{-}X=0
\end{equation}
for every \(X>0\).
\end{theorem}

We briefly describe the construction; the full proof and a concrete example for \(N=2\) are provided in the supplementary material.
Choose three scores \(a>b>c\) such that
\begin{equation}
    b-c>N \cdot (a-b).
\end{equation}
The \(N\) true matches receive score \(a\):
\begin{equation}
    S_{i,i}=a,
    \qquad
    i=0,\ldots,N-1.
\end{equation}
Next, assign score \(b\) to a false cycle that shifts each shared identity to the next one and uses the two open-set objects to close the cycle:
\begin{equation}
    A_0\rightarrow A_1,\quad
    A_1\rightarrow A_2,\quad
    \ldots,\quad
    A_{N-2}\rightarrow A_{N-1},\quad
    A_{N-1}\rightarrow U,\quad
    X\rightarrow A_0 .
\end{equation}
Equivalently,
\begin{equation}
    S_{i,i+1}=b\quad (i=0,\ldots,N-2),
    \qquad
    S_{N-1,N}=b,
    \qquad
    S_{N,0}=b.
\end{equation}
All remaining entries receive score \(c\). Since \(a>b\), every true pair is ranked above every false pair, which gives perfect AP and zero FPR-95. 
However, the correct assignment has score \(Na+c\), since it contains \(N\) true matches and one low-score open-set edge. 
The wrong cycle contains \(N+1\) false edges, each of score \(b\), and therefore has score \((N+1) \cdot b\). 
The condition above gives
\begin{equation}
    (N+1) \cdot b > N \cdot a+c,
\end{equation}
so Hungarian matching selects the all-false cycle.

\Cref{thm:perfect_ap_wrong_assignment} clarifies why AP and FPR-95 can disagree with Precision, Recall, Accuracy, and IPAA. Pairwise ranking metrics depend on the ordering of individual entries in the affinity matrix. Assignment-level metrics depend only on the final thresholded matching \(\mathcal{M}(S)\).

\subsection{Effect on Assignment-Level Metrics}
Unlike AP and FPR-95, assignment-level metrics depend on the thresholded Hungarian matching rather than on all pairwise scores. Sinkhorn normalization may therefore leave them unchanged when the selected matching is preserved, improve them by reinforcing the correct one-to-one structure, or degrade them by favoring an incorrect assignment or moving matches across the threshold. This motivates our empirical test of whether pairwise ranking metrics can improve without corresponding improvements in association quality.

\section{Post-processing Stress Test}
\label{sec:attack}
\begin{figure*}[t]
  \centering
  \includegraphics[width=0.9\linewidth]{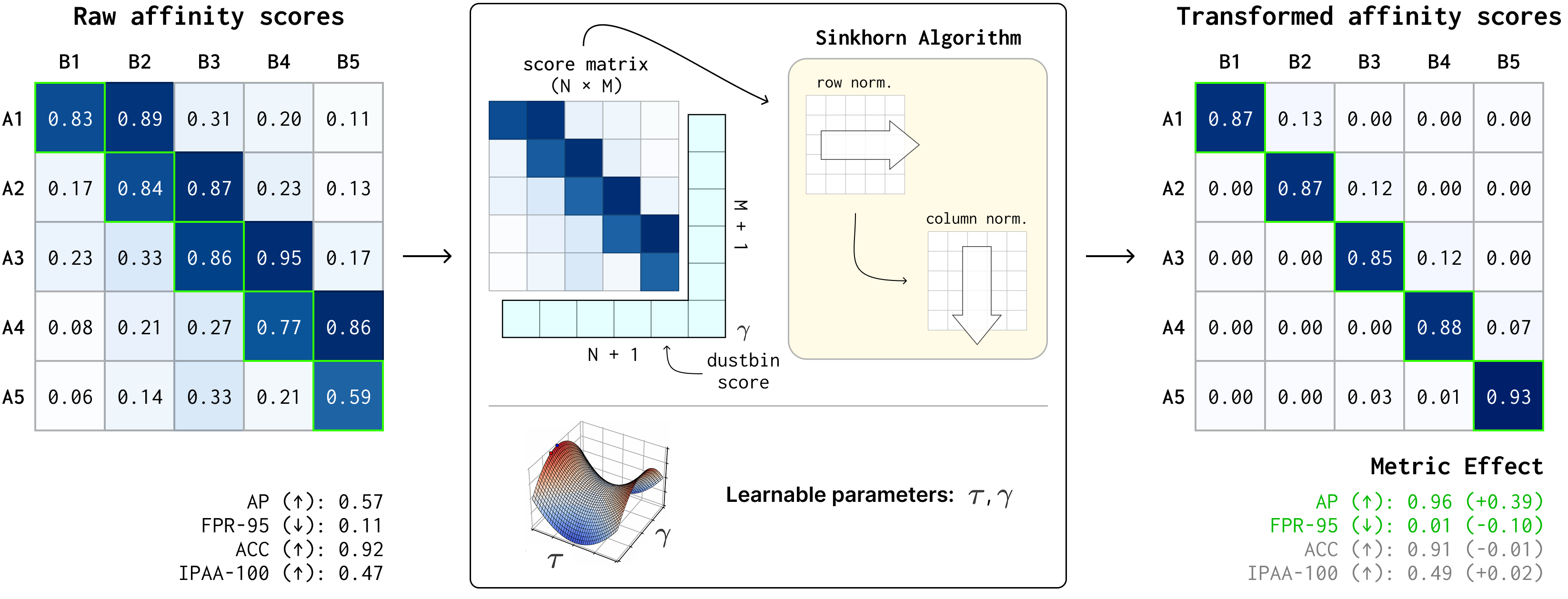}
  \caption{
    Overview of our Sinkhorn-based post-processing stress test.
    The raw affinity matrix is transformed by Sinkhorn normalization with learned temperature and dustbin parameters, producing scores that are more consistent with one-to-one matching.
    This can sharply improve pairwise ranking metrics without comparable gains in assignment-level metrics.
    }
  \label{fig:sinkhorn_attack}
\end{figure*}

Our theoretical analysis shows that pairwise ranking metrics can be sensitive to score transformations. We therefore introduce a Sinkhorn-based post-processing stress test that operates directly on the affinity matrices produced by a fixed association model. Our method does not modify detections, features, or model weights. Instead, it learns a small set of post-processing parameters that transform each affinity matrix before evaluation. The goal is to test whether AP and FPR-95 can be improved by reshaping the score matrix, without necessarily improving the final association decisions. \Cref{fig:sinkhorn_attack} illustrates the proposed pipeline.

\subsection{Dustbin Sinkhorn Transformation}
\label{sec:dustbin_sinkhorn}

Standard Sinkhorn normalization encourages all row and column mass to be assigned, which is restrictive in open-set multi-view association where detections may remain unmatched. 
To allow unmatched detections, we augment each affinity matrix with a dustbin row and column~(\Cref{fig:sinkhorn_attack}).
Given an affinity matrix \(S\in\mathbb{R}^{n\times m}\), we define
\begin{equation}
    S^{+}_{\gamma}
    =
    \begin{bmatrix}
        S & \gamma \cdot \mathbf{1}_{n} \\
        \gamma \cdot\mathbf{1}_{m}^{\top} & \gamma
    \end{bmatrix}
    \in \mathbb{R}^{(n+1)\times(m+1)},
\end{equation}
where \(\gamma\in\mathbb{R}\) is a learnable dustbin score. 
The augmented matrix is converted into a positive kernel and normalized with Sinkhorn:
\begin{equation}
    P^{+}_{\tau,\gamma}
    =
    \mathrm{Sinkhorn}_{K}
    \left(
        \exp(S^{+}_{\gamma}/\tau)
    \right),
\end{equation}
where \(\tau>0\) is a learnable temperature and \(K\) is the number of Sinkhorn iterations. In practice, Sinkhorn normalization is performed with non-uniform row and column marginals that allocate larger mass to the dustbin row and column. The exact marginals and log-space iterations are given in the supplementary material.
The stress test parameters are therefore
\begin{equation}
    \phi=(\tau,\gamma).
\end{equation}

After normalization, the real \(n\times m\) block of \(P^{+}_{\tau,\gamma}\) is used as the transformed cross-view affinity matrix. 
In implementation, this block is row-renormalized to keep the transformed values comparable to the original affinity scores used by thresholded Hungarian matching, as detailed in the supplementary material.
We denote the final transformed matrix as \(\widetilde{S}_{\phi}\).

\subsection{Optimization Objective}
\label{sec:attack_objective}

Our method optimizes only the post-processing parameters \(\phi=(\tau,\gamma)\). 
We use two loss terms. 
The first term encourages better pairwise ranking between positive and negative entries of the transformed matrix. 
Let \(\mathcal{P}\) and \(\mathcal{N}\) be the sets of positive and negative valid entries. 
We use a smooth \emph{pairwise ranking loss}:
\begin{equation}
    \mathcal{L}_{\mathrm{rank}}
    =
    \frac{1}{|\mathcal{P}||\mathcal{N}|}
    \sum_{p\in\mathcal{P}}
    \sum_{n\in\mathcal{N}}
    \log
    \left(
        1+
        \exp
        \left(
            -\frac{\widetilde{S}_{p}-\widetilde{S}_{n}}{\eta}
        \right)
    \right),
\end{equation}
where \(\eta>0\) is a ranking-loss temperature. 
This term pushes positive pairs above negative pairs and acts as a differentiable surrogate for improving AP and FPR-95.

The second term is an \emph{assignment drift loss} that discourages large changes in the original assignment structure. Let \(A(S)\in\{0,1\}^{n\times m}\) be the binary mask of the thresholded Hungarian matching computed from the original affinity matrix. We penalize deviations from this mask with a binary cross-entropy loss:
\begin{equation}
    \mathcal{L}_{\mathrm{drift}}
    =
    \mathrm{BCE}
    \left(
        \sigma(Z_{\phi}),
        A(S)
    \right),
\end{equation}
where \(Z_{\phi}\) denotes standardized logits obtained from the transformed scores. 
This term does not enforce identical assignments, but biases the optimization toward transformations that do not arbitrarily change the matching.

The final objective is
\begin{equation}
    \mathcal{L}
    =
    \mathcal{L}_{\mathrm{rank}}
    +
    \mathcal{L}_{\mathrm{drift}}.
\end{equation}
All affinity matrices are fixed during optimization, and gradients are applied only to \(\tau\) and \(\gamma\). 
The exact evaluation protocols and parameter selection criteria are described in \Cref{sec:experiments}.

%% file: sec/4_exps.tex
\section{Experiments}
\label{sec:experiments}

\subsection{Datasets and Association Methods}
\label{sec:datasets_methods}
We evaluate the proposed Sinkhorn-based post-processing stress test primarily on multi-view person association benchmarks, which provide a larger set of public models with different architectures, supervision regimes, and uses of geometry. To additionally test whether the observed effect extends beyond person association, we evaluate our stress test on the MessyTable~\cite{cai2020messytable} object association benchmark; these results are reported in the supplementary material.

We conduct experiments on WILDTRACK~\cite{chavdarova2018wildtrack} and MVOR~\cite{srivastav2018mvor}. Following the standard split used in prior multi-view association work~\cite{cheng2023rest,luna2022graph}, 70\% of the data is used for model training, 20\% for model validation, and 10\% for model testing. WILDTRACK is a multi-view video dataset captured by seven synchronized cameras in a public outdoor area with a dense group of pedestrians. MVOR is a multi-view operating-room dataset captured by three cameras and contains clinicians during surgical procedures, where similar clothing and occlusions make cross-view association challenging. 

We consider a diverse set of association methods, including appearance-based~\cite{zhou2021learning, mamedov2026retext}, geometry-aware~\cite{cai2020messytable}, graph-based~\cite{luna2022graph}, and self-supervised~\cite{gan2021self,chen2025learning} approaches. GNN-CCA is reported only on WILDTRACK because no official MVOR checkpoint was available. For each method and dataset split, we store the pairwise affinity matrices and apply our method without retraining any detector, feature extractor, or association model.
\subsection{Implementation Details}
\label{sec:implementation_details}
Our framework is implemented in PyTorch. All experiments are conducted on a desktop with an Intel Core i5-10600K CPU and an NVIDIA GeForce RTX 2060 GPU. For all evaluated association methods, we use the official implementations and publicly available checkpoints. The models are kept fixed throughout our experiments, and the stress test is applied only to the cached pairwise affinity matrices. We use \(K=20\) Sinkhorn iterations and optimize the post-processing parameters for 50 steps. Optimization is performed with full-batch gradient descent over all cached affinity matrices of the corresponding split, without mini-batching. We use the Adam optimizer~\cite{kingma2014adam} with learning rate \(0.03\). The temperature and dustbin score are initialized as \(\tau=0.015\) and \(\gamma=0.6\), respectively. The pairwise ranking loss temperature \(\eta\) is set to \(0.05\).

\subsection{Main Results}
\label{sec:main_results}

We evaluate the Sinkhorn-based post-processing stress test under two protocols.
In the \emph{test-to-test} protocol, the post-processing parameters are optimized directly on the test split and evaluated on the same split. This setting measures how much the reported metrics can be changed when the evaluation affinity matrices and labels are available.
In the \emph{val-to-test} protocol, the Sinkhorn parameters \((\tau,\gamma)\) are optimized on the validation split and then applied to the disjoint test split without using test labels. This is our main and more challenging protocol, since it tests whether the post-processing effect transfers to unseen test data without direct access to test annotations.

For each method, we compare the original affinity matrices with the Sinkhorn-transformed affinity matrices. Pairwise ranking metrics are computed from flattened affinity scores, while assignment-level metrics are computed after Hungarian matching and thresholding. For assignment-level metrics, we evaluate a fixed threshold grid and select the threshold \(\theta\) that maximizes ACC. In the \emph{test-to-test} protocol, \(\theta\) is selected on the test split. In the \emph{val-to-test} protocol, \(\theta\) is selected on the validation split and then fixed for test evaluation. Therefore, the absolute values of ACC and IPAA may slightly differ from those reported in the original papers. Additional metrics, including Precision and Recall, are reported in the supplementary material.

\begin{table*}[t]
\centering
\subfigure[Test-to-test protocol]{
\resizebox{\textwidth}{!}{
\begin{tabular}{llccccc}
\toprule
Dataset & Method
& AP \(\uparrow\)
& FPR-95 \(\downarrow\)
& ACC \(\uparrow\)
& IPAA-80 \(\uparrow\)
& IPAA-100 \(\uparrow\) \\
\midrule
\multirow{6}{*}{WILDTRACK}
& OSNet~\cite{zhou2021learning}
& \(0.167 \rightarrow \att{0.295}\)
& \(0.919 \rightarrow \att{0.838}\)
& \(0.438 \rightarrow \att{0.456}\)
& \(0.164 \rightarrow \att{0.148}\)
& \(0.004 \rightarrow \att{0.004}\) \\
& MvMHAT~\cite{gan2021self}
& \(0.237 \rightarrow \att{0.322}\)
& \(0.928 \rightarrow \att{0.820}\)
& \(0.439 \rightarrow \att{0.473}\)
& \(0.075 \rightarrow \att{0.095}\)
& \(0.004 \rightarrow \att{0.005}\) \\
& GNN-CCA~\cite{luna2022graph}
& \(0.493 \rightarrow \att{0.555}\)
& \(0.833 \rightarrow \att{0.420}\)
& \(0.618 \rightarrow \att{0.627}\)
& \(0.218 \rightarrow \att{0.231}\)
& \(0.008 \rightarrow \att{0.010}\) \\
& ReText~\cite{mamedov2026retext}
& \(0.528 \rightarrow \att{0.659}\)
& \(0.755 \rightarrow \att{0.457}\)
& \(0.659 \rightarrow \att{0.670}\)
& \(0.273 \rightarrow \att{0.300}\)
& \(0.020 \rightarrow \att{0.013}\) \\
& Self-MVA~\cite{chen2025learning}
& \(0.567 \rightarrow \att{0.955}\)
& \(0.112 \rightarrow \att{0.011}\)
& \(0.921 \rightarrow \att{0.911}\)
& \(0.901 \rightarrow \att{0.875}\)
& \(0.471 \rightarrow \att{0.485}\) \\
& OSNet + ESC~\cite{cai2020messytable}
& \(0.595 \rightarrow \att{0.810}\)
& \(0.127 \rightarrow \att{0.070}\)
& \(0.834 \rightarrow \att{0.828}\)
& \(0.687 \rightarrow \att{0.676}\)
& \(0.276 \rightarrow \att{0.286}\) \\
\midrule
\multirow{5}{*}{MVOR}
& MvMHAT~\cite{gan2021self}
& \(0.561 \rightarrow \att{0.802}\)
& \(0.832 \rightarrow \att{0.480}\)
& \(0.679 \rightarrow \att{0.670}\)
& \(0.470 \rightarrow \att{0.498}\)
& \(0.465 \rightarrow \att{0.493}\) \\
& OSNet~\cite{zhou2021learning}
& \(0.605 \rightarrow \att{0.797}\)
& \(0.801 \rightarrow \att{0.453}\)
& \(0.698 \rightarrow \att{0.679}\)
& \(0.488 \rightarrow \att{0.470}\)
& \(0.479 \rightarrow \att{0.460}\) \\
& ReText~\cite{mamedov2026retext}
& \(0.718 \rightarrow \att{0.889}\)
& \(0.629 \rightarrow \att{0.225}\)
& \(0.793 \rightarrow \att{0.784}\)
& \(0.662 \rightarrow \att{0.662}\)
& \(0.662 \rightarrow \att{0.662}\) \\
& OSNet + ESC~\cite{cai2020messytable}
& \(0.763 \rightarrow \att{0.888}\)
& \(0.399 \rightarrow \att{0.224}\)
& \(0.792 \rightarrow \att{0.792}\)
& \(0.709 \rightarrow \att{0.728}\)
& \(0.709 \rightarrow \att{0.728}\) \\
& Self-MVA~\cite{chen2025learning}
& \(0.840 \rightarrow \att{0.946}\)
& \(0.492 \rightarrow \att{0.103}\)
& \(0.836 \rightarrow \att{0.821}\)
& \(0.709 \rightarrow \att{0.714}\)
& \(0.704 \rightarrow \att{0.704}\) \\
\bottomrule
\end{tabular}
}
\label{tab:main_test_to_test}
}

\vspace{3pt}

\subfigure[Val-to-test protocol]{
\resizebox{\textwidth}{!}{
\begin{tabular}{llccccc}
\toprule
Dataset & Method
& AP \(\uparrow\)
& FPR-95 \(\downarrow\)
& ACC \(\uparrow\)
& IPAA-80 \(\uparrow\)
& IPAA-100 \(\uparrow\) \\
\midrule
\multirow{6}{*}{WILDTRACK}
& OSNet~\cite{zhou2021learning}
& \(0.167 \rightarrow \att{0.300}\)
& \(0.919 \rightarrow \att{0.779}\)
& \(0.429 \rightarrow \att{0.455}\)
& \(0.074 \rightarrow \att{0.105}\)
& \(0.004 \rightarrow \att{0.002}\) \\
& MvMHAT~\cite{gan2021self}
& \(0.237 \rightarrow \att{0.311}\)
& \(0.928 \rightarrow \att{0.779}\)
& \(0.437 \rightarrow \att{0.479}\)
& \(0.075 \rightarrow \att{0.104}\)
& \(0.004 \rightarrow \att{0.005}\) \\
& GNN-CCA~\cite{luna2022graph}
& \(0.493 \rightarrow \att{0.555}\)
& \(0.833 \rightarrow \att{0.447}\)
& \(0.615 \rightarrow \att{0.623}\)
& \(0.200 \rightarrow \att{0.214}\)
& \(0.008 \rightarrow \att{0.010}\) \\
& ReText~\cite{mamedov2026retext}
& \(0.528 \rightarrow \att{0.692}\)
& \(0.755 \rightarrow \att{0.376}\)
& \(0.659 \rightarrow \att{0.674}\)
& \(0.271 \rightarrow \att{0.295}\)
& \(0.020 \rightarrow \att{0.011}\) \\
& Self-MVA~\cite{chen2025learning}
& \(0.567 \rightarrow \att{0.953}\)
& \(0.112 \rightarrow \att{0.010}\)
& \(0.917 \rightarrow \att{0.901}\)
& \(0.867 \rightarrow \att{0.835}\)
& \(0.475 \rightarrow \att{0.463}\) \\
& OSNet + ESC~\cite{cai2020messytable}
& \(0.595 \rightarrow \att{0.809}\)
& \(0.127 \rightarrow \att{0.069}\)
& \(0.834 \rightarrow \att{0.824}\)
& \(0.687 \rightarrow \att{0.658}\)
& \(0.276 \rightarrow \att{0.279}\) \\
\midrule
\multirow{5}{*}{MVOR}
& MvMHAT~\cite{gan2021self}
& \(0.561 \rightarrow \att{0.799}\)
& \(0.832 \rightarrow \att{0.480}\)
& \(0.688 \rightarrow \att{0.676}\)
& \(0.521 \rightarrow \att{0.521}\)
& \(0.516 \rightarrow \att{0.516}\) \\
& OSNet~\cite{zhou2021learning}
& \(0.605 \rightarrow \att{0.797}\)
& \(0.801 \rightarrow \att{0.453}\)
& \(0.698 \rightarrow \att{0.664}\)
& \(0.435 \rightarrow \att{0.441}\)
& \(0.435 \rightarrow \att{0.441}\) \\
& ReText~\cite{mamedov2026retext}
& \(0.718 \rightarrow \att{0.894}\)
& \(0.629 \rightarrow \att{0.224}\)
& \(0.696 \rightarrow \att{0.661}\)
& \(0.676 \rightarrow \att{0.695}\)
& \(0.676 \rightarrow \att{0.695}\) \\
& OSNet + ESC~\cite{cai2020messytable}
& \(0.763 \rightarrow \att{0.884}\)
& \(0.399 \rightarrow \att{0.268}\)
& \(0.795 \rightarrow \att{0.790}\)
& \(0.714 \rightarrow \att{0.695}\)
& \(0.714 \rightarrow \att{0.695}\) \\
& Self-MVA~\cite{chen2025learning}
& \(0.840 \rightarrow \att{0.940}\)
& \(0.492 \rightarrow \att{0.175}\)
& \(0.830 \rightarrow \att{0.806}\)
& \(0.708 \rightarrow \att{0.685}\)
& \(0.703 \rightarrow \att{0.681}\) \\
\bottomrule
\end{tabular}
}
\label{tab:main_valid_to_test}
}
\caption{
Main results of our Sinkhorn-based post-processing stress test on WILDTRACK and MVOR.
Each cell reports the original score followed by the transformed score (\( \mathrm{raw}\rightarrow\mathrm{transformed} \)).
}
\label{tab:main_results}
\end{table*}

\begin{figure*}[t]
  \centering
  \includegraphics[width=0.6\linewidth]{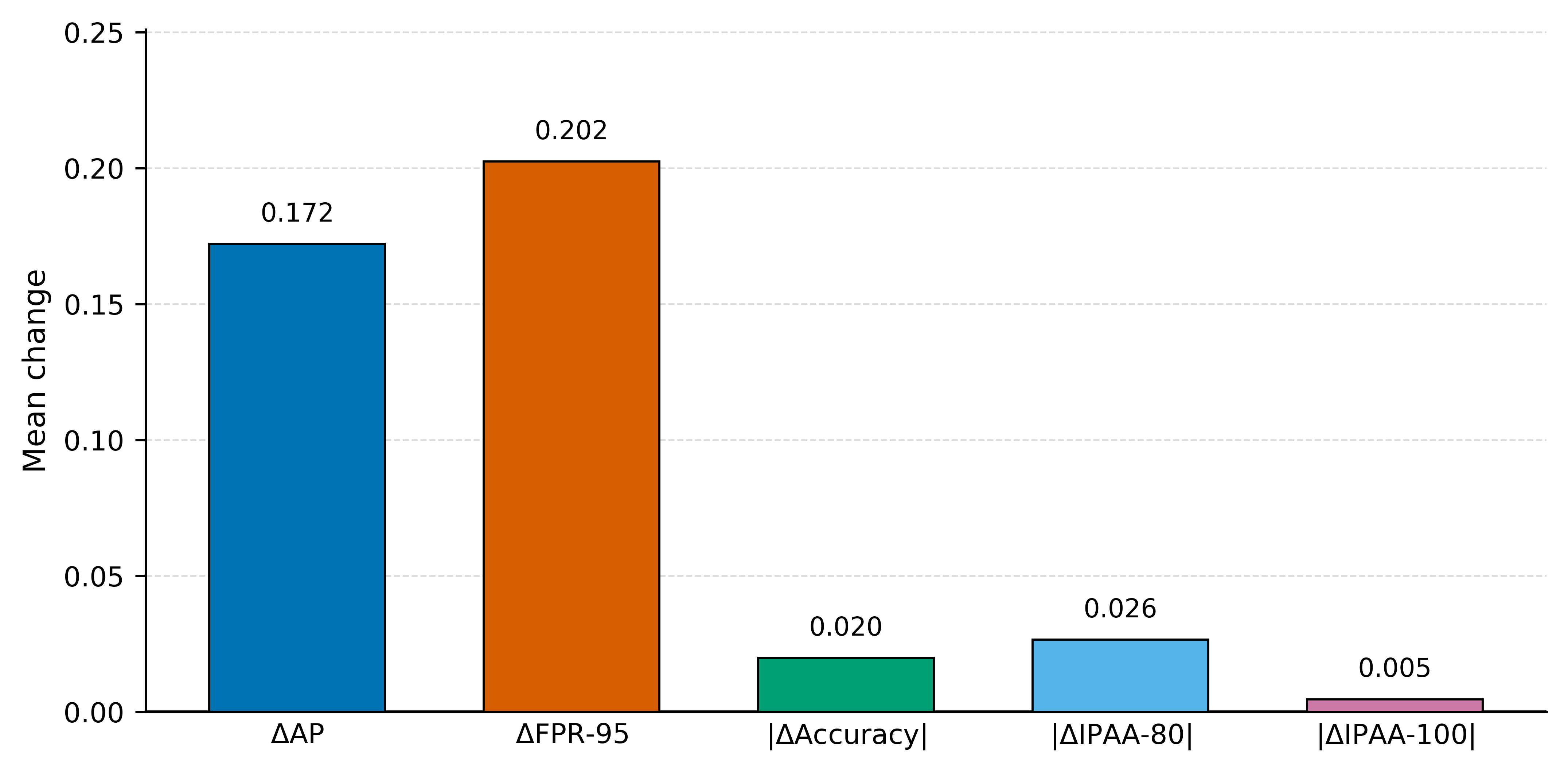}
  \caption{
Mean metric change across all evaluated methods on WILDTRACK under the validation-to-test protocol.
Our post-processing produces substantially larger changes in pairwise ranking metrics, AP and FPR-95, than in assignment-level metrics, ACC and IPAA.
}
\label{fig:metric_vulnerability}
\end{figure*}
\begin{figure*}[t]
  \centering
  \includegraphics[width=0.8\linewidth]{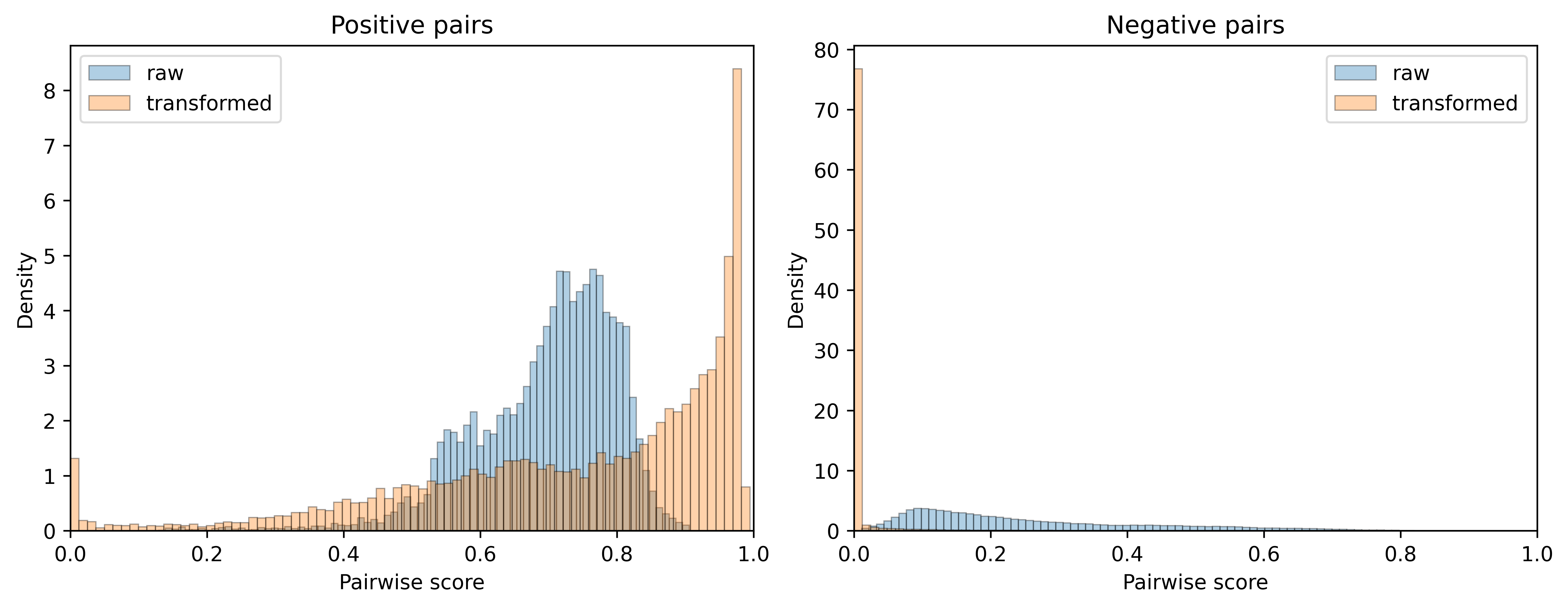}
  \caption{
Distribution of pairwise scores before and after the Sinkhorn-based post-processing.
}
\label{fig:score_distributions}
\end{figure*}
\begin{figure*}[t]
  \centering
  \includegraphics[width=0.6\linewidth]{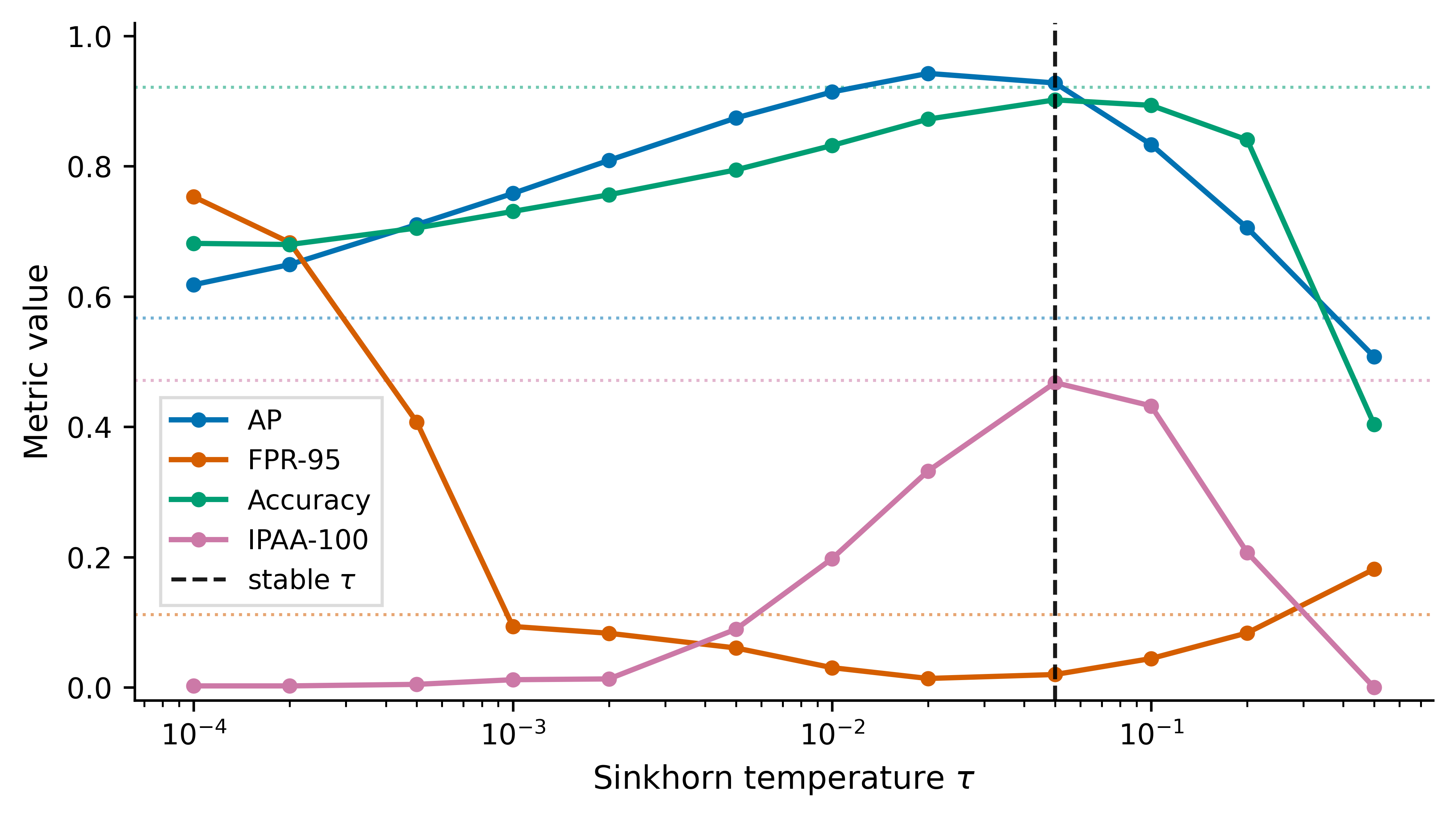}
  \caption{
Temperature sensitivity for Self-MVA on WILDTRACK under the test-to-test protocol.
Metrics are shown as a function of the Sinkhorn temperature \(\tau\); dotted horizontal lines denote the original values and the dashed vertical line marks the stable temperature.
}
\label{fig:attack_curves_tau}
\end{figure*}
\vspace{1pt}\noindent\textbf{Test-to-Test Results.}
\Cref{tab:main_results}(a) shows that Sinkhorn-based post-processing substantially improves pairwise ranking metrics on both datasets.
On WILDTRACK, the largest AP gain is observed for Self-MVA, from \(0.567\) to \(0.955\), while FPR-95 is consistently reduced, for example from \(0.833\) to \(0.420\) for GNN-CCA.
The same trend holds on MVOR, where ReText improves from \(0.718\) to \(0.889\) in AP and from \(0.629\) to \(0.225\) in FPR-95.
In contrast, ACC and IPAA usually remain close to their original values or move inconsistently. This suggests that improvements in pairwise ranking do not necessarily translate into better final associations.

\vspace{1pt}\noindent\textbf{Val-to-Test Results.}
\Cref{tab:main_results}(b) reports our main protocol, where both the post-processing parameters and the assignment threshold are selected on the validation split and then applied to the test split.
Despite not using test annotations for parameter or threshold selection, Sinkhorn-based post-processing still substantially improves AP and FPR-95 across methods and datasets.
On WILDTRACK, Self-MVA improves from \(0.567\) to \(0.953\) AP, while FPR-95 drops from \(0.112\) to \(0.010\).
On MVOR, MvMHAT improves from \(0.561\) to \(0.799\) AP, while FPR-95 decreases from \(0.832\) to \(0.480\).
These improvements are not accompanied by comparable gains in assignment-level metrics.
For example, on WILDTRACK, ReText changes from \(0.659\) to \(0.674\) in ACC and from \(0.271\) to \(0.295\) in IPAA-80.
On MVOR, OSNet + ESC changes from \(0.795\) to \(0.790\) in ACC and from \(0.714\) to \(0.695\) in IPAA-80.
This shows that AP and FPR-95 can still be strongly affected by validation-selected post-processing, while ACC and IPAA do not show corresponding improvements on the unseen test split. \Cref{fig:metric_vulnerability} summarizes this effect by averaging metric changes across different methods under the validation-to-test protocol on WILDTRACK. We further find that the Hungarian matching remains unchanged for the majority of affinity matrices after post-processing, reaching \(97.2\%\) unchanged matchings on MVOR (see supplementary material).



\vspace{1pt}\noindent\textbf{Score Distributions.}
\Cref{fig:score_distributions} illustrates how the Sinkhorn-based post-processing changes the distribution of pairwise scores. Positive pairs are shifted toward larger scores, with many of them concentrated near \(1\), while negative pairs are strongly compressed toward \(0\).
As a result, the overlap between positive and negative score distributions is greatly reduced.
This directly explains why AP increases and FPR-95 decreases after post-processing: our post-processing makes true pairs easier to rank above false pairs, even though this stronger pairwise separation does not necessarily lead to better final associations.

\vspace{1pt}\noindent\textbf{Temperature Sensitivity.} \Cref{fig:attack_curves_tau} shows the effect of the Sinkhorn temperature \(\tau\) for Self-MVA on WILDTRACK.
The curves reveal a non-monotonic trade-off between ranking and assignment quality.
Very small temperatures make the transformation too sharp and severely degrade ACC and IPAA, while larger temperatures gradually recover the assignment metrics.
At the selected stable point, \(\tau=0.05\), AP increases from \(0.567\) to \(0.928\) and FPR-95 decreases from \(0.112\) to \(0.020\), whereas ACC changes only from \(0.921\) to \(0.902\).
For even larger temperatures, the ranking gains start to decrease. In this regime, ACC and IPAA can also drop substantially, showing that Sinkhorn normalization can degrade the affinity matrix.

\subsection{Ablation Study}
\label{sec:ablation}
We conduct an ablation study to evaluate the main components of the Sinkhorn-based post-processing using Self-MVA on WILDTRACK as the default setting.

\vspace{2pt}\noindent\textbf{Dustbin Effect.}
\Cref{tab:ablation_dustbin} evaluates whether the dustbin component is necessary for our method.
Both variants optimize only the Sinkhorn temperature \(\tau\): the first applies Sinkhorn without a dustbin, while the second adds a dustbin with fixed score \(\gamma=0.6\).
Without the dustbin, Sinkhorn normalization improves AP and FPR-95, but degrades assignment-level metrics, reducing ACC from \(0.921\) to \(0.790\) and IPAA-100 from \(0.471\) to \(0.165\).
Adding the dustbin produces a much stronger effect on AP and FPR-95 and slightly improves ACC and IPAA-100 compared to the no-dustbin variant. This suggests that the dustbin helps stabilize the transformed assignment, but its strongest effect appears when combined with the full learnable post-processing setup.

\begin{table}[t]
\centering
\small
\fontsize{8pt}{10pt}\selectfont
\setlength{\tabcolsep}{3pt}
\renewcommand{\arraystretch}{0.95}

\begin{minipage}[t]{0.48\linewidth}
\centering
\begin{tabular}{c|cccc}
\hline
Dustbin & AP \(\uparrow\) & FPR-95 \(\downarrow\) & ACC \(\uparrow\) & IPAA-100 \(\uparrow\) \\
\hline
 & 0.767 & 0.142 & 0.790 & 0.165 \\
\rowcolor{gray!10}
\cmark & \textbf{0.803} & \textbf{0.117} & \textbf{0.798} & \textbf{0.175} \\
\hline
\end{tabular}
\caption{Effect of the dustbin component.}
\label{tab:ablation_dustbin}
\end{minipage}
\hfill
\begin{minipage}[t]{0.48\linewidth}
\centering
\begin{tabular}{cc|cccc}
\hline
\(\tau\) & \(\gamma\) & AP \(\uparrow\) & FPR-95 \(\downarrow\) & ACC \(\uparrow\) & IPAA-100 \(\uparrow\) \\
\hline
\cmark &  & 0.803 & 0.117 & 0.798 & 0.175 \\
 & \cmark & 0.946 & 0.018 & 0.904 & 0.435 \\
\rowcolor{gray!10}
\cmark & \cmark & \textbf{0.955} & \textbf{0.011} & \textbf{0.911} & \textbf{0.485} \\
\hline
\end{tabular}
\caption{Ablation of learnable post-processing parameters.}
\label{tab:ablation_params}
\end{minipage}

\end{table}
\begin{table}
\centering
\small
\fontsize{8pt}{10pt}\selectfont
\setlength{\tabcolsep}{4pt}
\renewcommand{\arraystretch}{0.95}
\begin{tabular}{cc|cccc}
\hline
\(\mathcal{L}_{\mathrm{drift}}\) & \(\mathcal{L}_{\mathrm{rank}}\) & AP \(\uparrow\) & FPR-95 \(\downarrow\) & ACC \(\uparrow\) & IPAA-100 \(\uparrow\) \\
\hline
\cmark &  & 0.774 & 0.150 & 0.765 & 0.174 \\
& \cmark & 0.944 & 0.020 & 0.906 & 0.432 \\
\rowcolor{gray!10}
\cmark & \cmark & \textbf{0.955} & \textbf{0.011} & \textbf{0.911} & \textbf{0.485} \\
\hline
\end{tabular}

\vspace{2pt}
\caption{
Ablation of our loss components.
}
\label{tab:ablation_losses}
\end{table}

\vspace{2pt}\noindent\textbf{Learnable Parameters.}
\Cref{tab:ablation_params} studies the effect of optimizing the Sinkhorn temperature \(\tau\), the dustbin score \(\gamma\), or both.
Optimizing only \(\tau\) with fixed \(\gamma = 0.6\) improves the ranking metrics but strongly harms assignment quality.
Optimizing only \(\gamma\) with fixed \(\tau = 0.015\) already achieves most of the stress test effect, while joint optimization gives the best overall result.
This suggests that the dustbin score is the dominant parameter in this setting, but tuning both parameters provides the most stable configuration.

\vspace{2pt}\noindent\textbf{Loss Components.}
\Cref{tab:ablation_losses} evaluates the two loss terms used to train our method.
The assignment drift loss alone provides no direct ranking signal and therefore gives only moderate AP/FPR-95 improvements while degrading assignment-level metrics.
The ranking loss is the main driver of the stress test, producing the strongest AP and FPR-95 improvements.
Adding the assignment drift loss slightly improves ACC and IPAA-100, indicating that it helps preserve the final assignment decisions without weakening the ranking metrics. We also compare our stress test against several lightweight score transformations, which produce substantially weaker effects on AP and FPR-95 (see supplementary material).

\section{Discussion and Recommendations}
\label{sec:discussion}
Our results suggest that AP and FPR-95 should be interpreted with caution in multi-view association. They remain useful diagnostics of pairwise affinity ranking, but should not be treated as sufficient evidence of association quality. Assignment-level metrics such as ACC and IPAA are better aligned with the task, since they evaluate the thresholded Hungarian matching. However, they remain local to individual image pairs and do not capture global consistency across all camera views. A promising direction is therefore to develop global multi-view metrics that evaluate the consistency of the full association graph and coherent multi-view identity groups.

%% file: sec/5_conclusion.tex
\section{Conclusion}
\label{sec:conclusion}
In this paper, we studied the mismatch between pairwise ranking metrics and constrained one-to-one matching in multi-view object association. Our theoretical analysis showed that this mismatch appears in two complementary ways: AP and FPR-95 can be imperfect even when the assignment is already correct, and Sinkhorn-based normalization can make them perfect. Conversely, optimal pairwise ranking can still lead to incorrect assignments. Empirically, we demonstrated this effect using a Sinkhorn-based post-processing stress test on multi-view person association benchmarks. Across multiple methods and datasets, AP and FPR-95 were substantially improved by tuning only a small number of post-processing parameters. Crucially, these improvements in pairwise ranking metrics are not reliably reflected in assignment-level metrics such as ACC and IPAA.

%% file: sec/6_supplementary.tex
\section{Pseudocode}
\label{app:pseudocode}
\addcontentsline{toc}{section}{Pseudocode}
For clarity, we provide the pseudocode for our Sinkhorn-based normalization~(\Cref{alg:dustbin_sinkhorn}).

\begin{algorithm}
\caption{Dustbin Sinkhorn normalization}
\label{alg:dustbin_sinkhorn}
\begin{algorithmic}[1]
\Require Affinity matrix \(S\in\mathbb{R}^{n\times m}\), temperature \(\tau\), dustbin score \(\gamma\), iterations \(K\)
\State Augment \(S\) with a dustbin row and column:
\[
S^{+}_{\gamma}
=
\begin{pmatrix}
S & \gamma \cdot \mathbf{1}_{n} \\
\gamma \cdot\mathbf{1}_{m}^{\top} & \gamma
\end{pmatrix}
\]
\State Set log-kernel \(Z=S^{+}_{\gamma}/\tau\)
\State Define row and column marginals:
\[
\mu=\frac{(1,\ldots,1,m)}{n+m},
\qquad
\nu=\frac{(1,\ldots,1,n)}{n+m}
\]
\State Initialize \(u=0,\ v=0\)
\For{\(k=1,\ldots,K\)}
    \State \(u \leftarrow \log \mu - \log\sum_j \exp(Z_{ij}+v_j)\)
    \State \(v \leftarrow \log \nu - \log\sum_i \exp(Z_{ij}+u_i)\)
\EndFor
\State \(\log P^{+}=Z+u\mathbf{1}^{\top}+\mathbf{1}v^{\top}\)
\State Extract real block \(P=P^{+}_{1:n,1:m}\)
\State Row-normalize the real block to obtain score-like affinities:
\[
\widetilde{S}_{ij}=P_{ij}/\mu_i
\]
\State \Return \(\widetilde{S}\)
\end{algorithmic}
\end{algorithm}

\section{Proof of Theorem 1}
\label{app:proof1}
\addcontentsline{toc}{section}{Proof of Theorem 1}
We prove the result by explicitly constructing the Sinkhorn-based transformation \(T_\tau\).
Let \(M^\star\in\mathcal{M}_k\) be the ground-truth matching, and define
\begin{equation}
    W^\star=\sum_{(i,j)\in M^\star} S_{ij},
    \qquad
    X^\star_{ij}=\mathbf{1}[(i,j)\in M^\star].
\end{equation}
For \(r=k+1,\ldots,\min(m,n)\), let
\begin{equation}
    \mathrm{OPT}_r
    =
    \max_{M\in\mathcal{M}_r}
    \sum_{(i,j)\in M} S_{ij}.
\end{equation}
If \(k<\min(m,n)\), define
\begin{equation}
    \Gamma
    =
    \max_{r=k+1,\ldots,\min(m,n)}
    \frac{\mathrm{OPT}_r-W^\star}{r-k},
\end{equation}
and if \(k=\min(m,n)\), set \(\Gamma=-\infty\).

We augment the rectangular matrix \(S\in\mathbb{R}^{m\times n}\) into a square matrix
\(\widetilde{S}\in\mathbb{R}^{N\times N}\), where
\begin{equation}
    N=m+n-k.
\end{equation}
Specifically, we add \(n-k\) dustbin rows and \(m-k\) dustbin columns. For a constant
\(C>\max\{0,\Gamma\}\), define
\begin{equation}
\widetilde{S}_{ij}
=
\begin{cases}
S_{ij}, & i\in[m],\ j\in[n],\\
0, & i\in[m],\ j\notin[n],\\
0, & i\notin[m],\ j\in[n],\\
-C, & i\notin[m],\ j\notin[n].
\end{cases}
\end{equation}
Thus, mixed real-dustbin assignments have zero score, while dustbin-dustbin assignments receive a penalty \(-C\).

Let
\begin{equation}
    \mathcal{B}_N
    =
    \{P\in\mathbb{R}_{\geq0}^{N\times N}: P\mathbf{1}=\mathbf{1},\ P^\top\mathbf{1}=\mathbf{1}\}
\end{equation}
be the Birkhoff polytope. We first show that every optimum of the linear assignment problem over \(\mathcal{B}_N\) has the same real block \(X^\star\).

Consider a permutation \(\pi\) of \([N]\) and its permutation matrix \(P_\pi\). Let
\begin{equation}
    M(\pi)=\{(i,j)\in[m]\times[n]: \pi(i)=j\}
\end{equation}
be the real matching induced by \(\pi\), and let \(r=|M(\pi)|\). Since \(r\) real rows are assigned to real columns, \(m-r\) real rows are assigned to dustbin columns. There are \(m-k\) dustbin columns in total, hence the number of dustbin columns assigned to dustbin rows is
\begin{equation}
    (m-k)-(m-r)=r-k.
\end{equation}
Therefore \(r\geq k\), and the number of dustbin-dustbin assignments is exactly \(r-k\). By construction of \(\widetilde{S}\),
\begin{equation}
    \langle P_\pi,\widetilde{S}\rangle
    =
    \sum_{(i,j)\in M(\pi)} S_{ij}
    -
    C\cdot (r-k).
\end{equation}

There exists a permutation inducing exactly \(M^\star\) in the real block: unmatched real rows are assigned to dustbin columns and dustbin rows are assigned to unmatched real columns. For this permutation \(r=k\), there are no dustbin-dustbin assignments, and the total score is \(W^\star\).

Now consider any permutation \(\pi\). If \(r=k\), then
\begin{equation}
    \langle P_\pi,\widetilde{S}\rangle
    =
    \sum_{(i,j)\in M(\pi)} S_{ij}.
\end{equation}
By the uniqueness assumption on \(M^\star\), this value equals \(W^\star\) only when
\(M(\pi)=M^\star\); otherwise it is strictly smaller.

If \(r>k\), then
\begin{equation}
    \langle P_\pi,\widetilde{S}\rangle
    \leq
    \mathrm{OPT}_r-C \cdot (r-k).
\end{equation}
Since \(C>\Gamma\), we have
\begin{equation}
    C>
    \frac{\mathrm{OPT}_r-W^\star}{r-k},
\end{equation}
and therefore
\begin{equation}
    \mathrm{OPT}_r-C \cdot (r-k)<W^\star.
\end{equation}
Hence every permutation with \(r>k\) has score strictly smaller than \(W^\star\).

Thus, the optimal permutation matrices of the augmented linear assignment problem are exactly those whose real block is \(X^\star\). Since a linear functional over \(\mathcal{B}_N\) attains its maximum on the convex hull of its optimal vertices, every optimal solution \(P\) of
\begin{equation}
    \max_{P\in\mathcal{B}_N}\langle P,\widetilde{S}\rangle
\end{equation}
satisfies
\begin{equation}
    P|_{[m]\times[n]}=X^\star.
\end{equation}

We now define the Sinkhorn-based transformation. For \(\tau>0\), let
\begin{equation}
    P_\tau
    =
    \arg\max_{P\in\mathcal{B}_N}
    \left\{
        \langle P,\widetilde{S}\rangle
        +
        \tau \cdot H(P)
    \right\},
\end{equation}
where
\begin{equation}
    H(P)=-\sum_{i,j}P_{ij} \cdot \log P_{ij},
\end{equation}
with the convention \(0\log 0=0\). This entropic assignment solution is equivalently obtained by Sinkhorn scaling of \(\exp(\widetilde{S}/\tau)\). We define
\begin{equation}
    T_\tau(S)=P_\tau|_{[m]\times[n]}.
\end{equation}

It remains to show that \(T_\tau(S)\to X^\star\). Let
\begin{equation}
    L(P)=\langle P,\widetilde{S}\rangle,
    \qquad
    L^\star=\max_{P\in\mathcal{B}_N}L(P).
\end{equation}
Let \(Q^\star\) be any optimizer of the unregularized linear problem. By optimality of \(P_\tau\),
\begin{equation}
    L(P_\tau)+\tau \cdot H(P_\tau)
    \geq
    L(Q^\star)+\tau \cdot H(Q^\star)
    =
    L^\star+\tau \cdot H(Q^\star).
\end{equation}
Since \(L(P_\tau)\leq L^\star\), we get
\begin{equation}
    0
    \leq
    L^\star-L(P_\tau)
    \leq
    \tau\bigl(H(P_\tau)-H(Q^\star)\bigr).
\end{equation}
The entropy is bounded on \(\mathcal{B}_N\), for example \(0\leq H(P)\leq N\log N\). Hence
\begin{equation}
    L(P_\tau)\rightarrow L^\star
    \qquad
    \text{as}
    \qquad
    \tau\rightarrow0^+.
\end{equation}

Take any sequence \(\tau_\ell\to0^+\). By compactness of \(\mathcal{B}_N\), a subsequence of \(P_{\tau_\ell}\) converges to some \(Q\in\mathcal{B}_N\). By continuity of \(L\), the limit satisfies \(L(Q)=L^\star\), so \(Q\) is an optimal solution of the unregularized linear problem. From the argument above, every such optimizer has real block \(X^\star\). Therefore,
\begin{equation}
    Q|_{[m]\times[n]}=X^\star.
\end{equation}
Since every limit point has the same real block, the entire family satisfies
\begin{equation}
    T_\tau(S)=P_\tau|_{[m]\times[n]}
    \rightarrow X^\star
    \qquad
    \text{as}
    \qquad
    \tau\rightarrow0^+.
\end{equation}

Finally, because the number of real entries is finite, this convergence implies that there exists \(\tau_0>0\) such that for all \(0<\tau<\tau_0\),
\begin{equation}
    T_\tau(S)_{ij}>\frac{2}{3}
    \quad \text{for all } (i,j)\in M^\star,
    \qquad
    T_\tau(S)_{ij}<\frac{1}{3}
    \quad \text{for all } (i,j)\notin M^\star.
\end{equation}
Thus all positive pairs are ranked strictly above all negative pairs, and therefore
\begin{equation}
    \mathrm{AP}(T_\tau(S))=1.
\end{equation}
If \(mn-k\geq1\), then there is at least one negative pair, and the threshold \(t=1/2\) accepts all positives and rejects all negatives. Hence
\begin{equation}
    \mathrm{TPR}(t)=1,
    \qquad
    \mathrm{FPR}(t)=0.
\end{equation}
Since \(\mathrm{TPR}(t)\geq0.95\), the definition of FPR-95 gives
\begin{equation}
    \mathrm{FPR}\text{-}95(T_\tau(S))=0.
\end{equation}
This completes the proof.

\vspace{1pt}\noindent\textbf{Nontrivial Example.} The assumption above does not imply that the original pairwise ranking is perfect. 
For example, let \(m=2\), \(n=3\), \(k=2\), \(M^\star=\{(1,1),(2,2)\}\), and
\begin{equation}
S=
\begin{pmatrix}
0.49 & 0.50 & 0.20\\
0.44 & 0.48 & 0.22
\end{pmatrix}.
\end{equation}
The ground-truth matching has score \(0.49+0.48=0.97\), while every other matching of size \(2\) has smaller total score. Thus the constrained assignment recovers \(M^\star\). 
However, the negative pair \((1,2)\) has score \(0.50\), which is larger than the positive pair \((2,2)\) with score \(0.48\). 
Therefore, the original pairwise ranking is not perfect, so AP is below \(1\) and FPR-95 is above \(0\). 
The theorem shows that the Sinkhorn-based transformation can make these pairwise metrics perfect while the constrained assignment was already correct.

\section{Proof of Theorem 2}
\label{app:proof2}
\addcontentsline{toc}{section}{Proof of Theorem 2}
We prove the claim by construction. 
Let the two views contain
\[
A_0,\ldots,A_{N-1},X
\quad\text{and}\quad
A_0,\ldots,A_{N-1},U,
\]
respectively. 
The shared identities are \(A_0,\ldots,A_{N-1}\), so the true matches are
\[
(A_i,A_i), \qquad i=0,\ldots,N-1.
\]
The objects \(X\) and \(U\) are open-set objects and should remain unmatched.

Choose three scores \(a>b>c\) such that
\begin{equation}
    b-c > N \cdot (a-b).
    \label{eq:wrong_cycle_condition}
\end{equation}
For example, one can take \(a=1\), \(b=1-\varepsilon\), and \(c=0\) for any
\(0<\varepsilon<1/(N+1)\).

We construct \(S\in\mathbb{R}^{(N+1)\times(N+1)}\) as follows. 
The \(N\) true pairs receive score \(a\):
\begin{equation}
    S_{i,i}=a,
    \qquad
    i=0,\ldots,N-1.
\end{equation}
Next, assign score \(b\) to the following false cycle:
\begin{equation}
    A_0\rightarrow A_1,\quad
    A_1\rightarrow A_2,\quad
    \ldots,\quad
    A_{N-2}\rightarrow A_{N-1},\quad
    A_{N-1}\rightarrow U,\quad
    X\rightarrow A_0 .
\end{equation}
Equivalently,
\begin{equation}
    S_{i,i+1}=b\quad (i=0,\ldots,N-2),
    \qquad
    S_{N-1,N}=b,
    \qquad
    S_{N,0}=b.
\end{equation}
All remaining entries are assigned score \(c\).

First, every true pair has score \(a\), while every false pair has score at most \(b\). 
Since \(a>b\), all positives are ranked above all negatives in the flattened list of pairwise scores. 
Therefore,
\begin{equation}
    \mathrm{AP}(S)=1.
\end{equation}
Moreover, any threshold \(t\in(b,a]\) accepts all true pairs and rejects all false pairs, so \(\mathrm{TPR}(t)=1\) and \(\mathrm{FPR}(t)=0\). 
Hence,
\begin{equation}
    \mathrm{FPR}\text{-}95(S)=0.
\end{equation}

It remains to show that Hungarian matching selects the false cycle. 
The assignment that matches all shared identities correctly and assigns \(X\) to \(U\) has total score
\begin{equation}
    N \cdot a+c.
\end{equation}
The false cycle assignment contains \(N+1\) false edges, each with score \(b\), and therefore has total score
\begin{equation}
    (N+1) \cdot b.
\end{equation}
Condition~\eqref{eq:wrong_cycle_condition} is equivalent to
\begin{equation}
    (N+1) \cdot b > N \cdot a+c.
\end{equation}
Thus, the false cycle is better than the natural assignment that contains all true matches.

We now show that the false cycle is globally optimal. 
Consider any permutation assignment.

If the assignment uses no true diagonal edge and is not the false cycle, then it cannot use all \(N+1\) cycle edges. 
Therefore, it uses at most \(N\) edges of score \(b\) and at least one edge of score \(c\). 
Its total score is at most
\begin{equation}
    N \cdot b+c < (N+1) \cdot b,
\end{equation}
so it is worse than the false cycle.

Now suppose the assignment uses \(q\geq1\) true diagonal edges. 
Once a true diagonal edge is used, the false cycle is broken, so the assignment must include at least one edge outside the true diagonals and outside the high-score cycle; such an edge has score \(c\). 
The best possible total score of such an assignment is therefore at most
\begin{equation}
    q \cdot a + (N-q) \cdot b + c.
\end{equation}
Indeed, the remaining \(N-q\) non-\(c\) edges can have score at most \(b\), and one additional edge has score \(c\). 
Rewriting this upper bound gives
\begin{equation}
    q \cdot a + (N-q) \cdot b + c
    =
    (N+1) \cdot b + q \cdot (a-b) - (b-c).
\end{equation}
Since \(q\leq N\), we have
\begin{equation}
    q\cdot (a-b) \leq N \cdot (a-b) < b-c
\end{equation}
by \eqref{eq:wrong_cycle_condition}. Hence
\begin{equation}
    q \cdot a + (N-q) \cdot b + c < (N+1) \cdot b.
\end{equation}
Thus, every assignment using at least one true match is also worse than the false cycle. 
Therefore, Hungarian matching selects the false cycle, and all selected matches are false.

Finally, choose any threshold \(\theta\) such that
\begin{equation}
    c < \theta \leq b.
\end{equation}
All edges of the false cycle have score \(b\), so they remain after thresholding, while low-score \(c\) edges are discarded. 
The retained predicted matches are exactly the false cycle. 
Thus the number of correct predicted matches is zero, while \(N+1\) matches are predicted:
\begin{equation}
    \mathrm{Precision}=\frac{0}{N+1}=0.
\end{equation}
Since none of the \(N\) ground-truth cross-view matches is recovered,
\begin{equation}
    \mathrm{Recall}=\frac{0}{N}=0.
\end{equation}
For the association accuracy used in the main paper, the numerator counts correct predicted matches and correctly unmatched identities. 
Here, there are no correct predicted matches. 
Moreover, neither open-set object is correctly left unmatched: \(X\) is matched to \(A_0\), and \(U\) is matched from \(A_{N-1}\). 
Hence the number of correctly unmatched identities is also zero, and therefore
\begin{equation}
    \mathrm{ACC}=0.
\end{equation}
For each image pair, IPAA-\(X\) is the indicator that the association accuracy is at least \(X/100\). 
Since the association accuracy is zero, we have
\begin{equation}
    \mathrm{IPAA}\text{-}X=0
\end{equation}
for every \(X>0\). 
This completes the proof.

\vspace{1pt}\noindent\textbf{Nontrivial Example.} For \(N=2\), one possible matrix is
\begin{equation}
S =
\begin{pmatrix}
0.90 & 0.89 & 0.01 \\
0.01 & 0.90 & 0.89 \\
0.89 & 0.01 & 0.01
\end{pmatrix},
\end{equation}
where rows are ordered as \((A_0,A_1,X)\) and columns as \((A_0,A_1,U)\). Both true pairs have score \(0.90\), while all false pairs have score at most \(0.89\), so the pairwise ranking is perfect.

However, Hungarian matching maximizes the total assignment score and therefore selects the higher-scoring false cycle
\[
    A_0\rightarrow A_1,
    \quad
    A_1\rightarrow U,
    \quad
    X\rightarrow A_0 .
\]
This cycle has total score \(3\cdot 0.89=2.67\), which is larger than the assignment containing the two true matches and the open-set edge,
\[
    A_0\rightarrow A_0,
    \quad
    A_1\rightarrow A_1,
    \quad
    X\rightarrow U,
\]
whose total score is \(0.90+0.90+0.01=1.81\). Thus, the pairwise metrics report perfect separability, while the induced assignment is completely wrong. 

\section{Additional Precision and Recall Results}
\label{app:precision_recall}
\addcontentsline{toc}{section}{Additional Precision and Recall Results}
In the main paper, we focus on AP, FPR-95, ACC, and IPAA. 
For completeness, we additionally report Precision and Recall under the test-to-test protocol in \Cref{tab:supp_precision_recall}. Precision and Recall change less systematically than AP and FPR-95. 
This is expected because they are computed after Hungarian matching and thresholding, where the threshold \(\theta\) induces a trade-off between the two metrics. 
A stricter threshold can increase Precision by discarding uncertain matches, while a smaller threshold can increase Recall by keeping more associations. 
Since the threshold is re-selected after Sinkhorn-based post-processing, changes in Precision and Recall reflect both the transformed scores and the newly selected threshold.

\begin{table*}[t]
\centering
\subfigure[WILDTRACK]{
\resizebox{0.5\textwidth}{!}{
\begin{tabular}{lcc}
\toprule
Method & Precision \(\uparrow\) & Recall \(\uparrow\) \\
\midrule
OSNet~\cite{zhou2021learning}
& \(0.511 \rightarrow \att{0.644}\)
& \(0.168 \rightarrow \att{0.183}\) \\
MvMHAT~\cite{gan2021self}
& \(0.349 \rightarrow \att{0.489}\)
& \(0.357 \rightarrow \att{0.320}\) \\
GNN-CCA~\cite{luna2022graph}
& \(0.987 \rightarrow \att{0.964}\)
& \(0.397 \rightarrow \att{0.415}\) \\
ReText~\cite{mamedov2026retext}
& \(0.694 \rightarrow \att{0.829}\)
& \(0.582 \rightarrow \att{0.524}\) \\
OSNet + ESC~\cite{cai2020messytable}
& \(0.799 \rightarrow \att{0.798}\)
& \(0.803 \rightarrow \att{0.799}\) \\
Self-MVA~\cite{chen2025learning}
& \(0.938 \rightarrow \att{0.919}\)
& \(0.928 \rightarrow \att{0.930}\) \\
\bottomrule
\end{tabular}
}
\label{tab:supp_precision_recall_wildtrack}
}
\hfill
\subfigure[MVOR]{
\resizebox{0.5\textwidth}{!}{
\begin{tabular}{lcc}
\toprule
Method & Precision \(\uparrow\) & Recall \(\uparrow\) \\
\midrule
MvMHAT~\cite{gan2021self}
& \(0.777 \rightarrow \att{0.754}\)
& \(0.688 \rightarrow \att{0.693}\) \\
OSNet~\cite{zhou2021learning}
& \(0.794 \rightarrow \att{0.783}\)
& \(0.699 \rightarrow \att{0.682}\) \\
ReText~\cite{mamedov2026retext}
& \(0.842 \rightarrow \att{0.832}\)
& \(0.803 \rightarrow \att{0.795}\) \\
OSNet + ESC~\cite{cai2020messytable}
& \(0.816 \rightarrow \att{0.813}\)
& \(0.822 \rightarrow \att{0.838}\) \\
Self-MVA~\cite{chen2025learning}
& \(0.910 \rightarrow \att{0.903}\)
& \(0.893 \rightarrow \att{0.889}\) \\
\bottomrule
\end{tabular}
}
\label{tab:supp_precision_recall_mvor}
}
\caption{
Additional Precision and Recall results under the test-to-test protocol.
Each cell reports the original value followed by the value after the Sinkhorn-based post-processing stress test
(\(\mathrm{raw}\rightarrow\mathrm{transformed}\)).
}
\label{tab:supp_precision_recall}
\end{table*}

\section{Grid Search over Sinkhorn Parameters}
\label{app:grid_search}
\addcontentsline{toc}{section}{Grid Search over Sinkhorn Parameters}

In the main experiments, the Sinkhorn temperature \(\tau\) and dustbin score \(\gamma\) are optimized by gradient descent. As an additional baseline, we evaluate a simple grid search over these two parameters on WILDTRACK under the test-to-test protocol (\Cref{tab:supp_grid_search}). We use the following grid:
\begin{equation}
\begin{split}
\tau \in \{&
0.0001,0.0002,0.0005,0.001,0.002,0.005,0.007,0.01,\\
&0.02,0.03,0.05,0.1,0.2,0.3,0.5\},
\end{split}
\end{equation}
\begin{equation}
\gamma \in
\{-0.5,-0.25,0.0,0.05,0.1,0.2,0.3,0.4,0.5,0.75,1.0\}.
\end{equation}

\begin{table*}[t]
\centering
\small
\setlength{\tabcolsep}{3.5pt}
\renewcommand{\arraystretch}{0.95}
\resizebox{\textwidth}{!}{
\begin{tabular}{l|ccccc}
\toprule
Method
& AP \(\uparrow\)
& FPR-95 \(\downarrow\)
& ACC \(\uparrow\)
& IPAA-80 \(\uparrow\)
& IPAA-100 \(\uparrow\) \\
\midrule
OSNet~\cite{zhou2021learning}
& \(0.167 \rightarrow \att{0.279}\)
& \(0.919 \rightarrow \att{0.842}\)
& \(0.438 \rightarrow \att{0.447}\)
& \(0.164 \rightarrow \att{0.125}\)
& \(0.004 \rightarrow \att{0.002}\) \\

MvMHAT~\cite{gan2021self}
& \(0.237 \rightarrow \att{0.251}\)
& \(0.928 \rightarrow \att{0.819}\)
& \(0.439 \rightarrow \att{0.444}\)
& \(0.075 \rightarrow \att{0.067}\)
& \(0.004 \rightarrow \att{0.000}\) \\

GNN-CCA~\cite{luna2022graph}
& \(0.493 \rightarrow \att{0.522}\)
& \(0.833 \rightarrow \att{0.610}\)
& \(0.618 \rightarrow \att{0.621}\)
& \(0.218 \rightarrow \att{0.200}\)
& \(0.008 \rightarrow \att{0.012}\) \\

ReText~\cite{mamedov2026retext}
& \(0.528 \rightarrow \att{0.628}\)
& \(0.755 \rightarrow \att{0.520}\)
& \(0.659 \rightarrow \att{0.661}\)
& \(0.273 \rightarrow \att{0.269}\)
& \(0.020 \rightarrow \att{0.014}\) \\

Self-MVA~\cite{chen2025learning}
& \(0.567 \rightarrow \att{0.926}\)
& \(0.112 \rightarrow \att{0.021}\)
& \(0.921 \rightarrow \att{0.908}\)
& \(0.901 \rightarrow \att{0.845}\)
& \(0.471 \rightarrow \att{0.483}\) \\

OSNet + ESC~\cite{cai2020messytable}
& \(0.595 \rightarrow \att{0.833}\)
& \(0.127 \rightarrow \att{0.059}\)
& \(0.834 \rightarrow \att{0.810}\)
& \(0.687 \rightarrow \att{0.619}\)
& \(0.276 \rightarrow \att{0.238}\) \\
\bottomrule
\end{tabular}
}
\caption{
Grid search over Sinkhorn temperature \(\tau\) and dustbin score \(\gamma\) on WILDTRACK under the test-to-test protocol.
Each cell reports the original value followed by the best grid-search value
(\(\mathrm{raw}\rightarrow\mathrm{grid}\)).
}
\label{tab:supp_grid_search}
\end{table*}

Grid search improves AP and FPR-95 for all methods, confirming that the metric mismatch can be exposed even without gradient-based optimization.
However, the gains are generally weaker than those obtained by the learned Sinkhorn-based post-processing in the main paper.
For example, for Self-MVA, grid search improves AP from \(0.567\) to \(0.926\), whereas the learned method reaches \(0.955\).
For GNN-CCA, grid search reduces FPR-95 from \(0.833\) to \(0.610\), while gradient-based optimization reduces it to \(0.420\).
A similar gap appears for ReText, where grid search reaches \(0.628\) AP and \(0.520\) FPR-95, compared to \(0.659\) AP and \(0.457\) FPR-95 with the learned post-processing.
These results indicate that the two-parameter Sinkhorn transformation is already sensitive under coarse parameter search, but gradient optimization provides a stronger and more reliable way to identify configurations that substantially affect the pairwise ranking metrics.

\section{Comparison with Simple Post-processing Baselines}
\label{app:simple_baselines}

We also compare our Sinkhorn-based post-processing stress test with three lightweight post-processing baselines on Self-MVA under the WILDTRACK test-to-test protocol~(\Cref{tab:supp_simple_baselines}).
Given an affinity matrix \(S\), row softmax is defined as
\begin{equation}
    S'_{ij}
    =
    \frac{\exp(S_{ij}/\tau)}
    {\sum_{j'} \exp(S_{ij'}/\tau)}.
\end{equation}
Row min-max normalization rescales each row independently:
\begin{equation}
    S'_{ij}
    =
    \frac{S_{ij}-\min_{j'}S_{ij'}}
    {\max_{j'}S_{ij'}-\min_{j'}S_{ij'}+\epsilon}.
\end{equation}
Finally, row+column softmax averages row-wise and column-wise softmax scores:
\begin{equation}
    S'
    =
    \frac{1}{2}
    \left(
    \mathrm{RowSoftmax}(S/\tau)
    +
    \mathrm{ColSoftmax}(S/\tau)
    \right).
\end{equation}

\begin{table*}[t]
\centering
\small
\setlength{\tabcolsep}{3.5pt}
\renewcommand{\arraystretch}{0.95}
\resizebox{\textwidth}{!}{
\begin{tabular}{lccccc}
\toprule
Post-processing
& AP \(\uparrow\)
& FPR-95 \(\downarrow\)
& ACC \(\uparrow\)
& IPAA-80 \(\uparrow\)
& IPAA-100 \(\uparrow\) \\
\midrule
Row min-max
& \(0.567 \rightarrow 0.482\)
& \(0.112 \rightarrow 0.078\)
& \(0.921 \rightarrow 0.808\)
& \(0.901 \rightarrow 0.520\)
& \(0.471 \rightarrow 0.231\) \\

Row softmax
& \(0.567 \rightarrow 0.489\)
& \(0.112 \rightarrow 0.116\)
& \(0.921 \rightarrow 0.858\)
& \(0.901 \rightarrow 0.746\)
& \(0.471 \rightarrow 0.300\) \\

Row+column softmax
& \(0.567 \rightarrow 0.611\)
& \(0.112 \rightarrow 0.098\)
& \(0.921 \rightarrow 0.867\)
& \(0.901 \rightarrow 0.758\)
& \(0.471 \rightarrow 0.355\) \\

\rowcolor{gray!10}
Ours
& \(0.567 \rightarrow \att{0.955}\)
& \(0.112 \rightarrow \att{0.011}\)
& \(0.921 \rightarrow \att{0.911}\)
& \(0.901 \rightarrow \att{0.875}\)
& \(0.471 \rightarrow \att{0.485}\) \\
\bottomrule
\end{tabular}
}
\caption{
Comparison with simple post-processing baselines on Self-MVA under the WILDTRACK test-to-test protocol.
Each cell reports the original value followed by the post-processed value
(\(\mathrm{raw}\rightarrow\mathrm{processed}\)).
}
\label{tab:supp_simple_baselines}
\end{table*}

Simple post-processing baselines do not reproduce the behavior of our Sinkhorn-based post-processing. In contrast, our method improves AP to \(0.955\) and FPR-95 to \(0.011\), while keeping ACC and IPAA much closer to their original values. These results suggest that the observed metric sensitivity is not merely a consequence of row-wise calibration or independent score rescaling. The dustbin Sinkhorn transformation is more effective because it reshapes the matrix using a global one-to-one structure while still allowing unmatched detections through the dustbin.

\section{Additional Results on MessyTable}
\label{app:messytable_benchmark}

To examine whether the observed metric sensitivity is specific to person association, we additionally evaluate our Sinkhorn-based post-processing stress test on the MessyTable~\cite{cai2020messytable} object association benchmark. We evaluate both the test-to-test and val-to-test protocols~(\Cref{tab:supp_messytable}).

\begin{table*}[t]
\centering
\small
\setlength{\tabcolsep}{3.5pt}
\renewcommand{\arraystretch}{0.95}
\resizebox{\textwidth}{!}{
\begin{tabular}{llccccc}
\toprule
Protocol & Method
& AP \(\uparrow\)
& FPR-95 \(\downarrow\)
& ACC \(\uparrow\)
& IPAA-80 \(\uparrow\)
& IPAA-100 \(\uparrow\) \\
\midrule
\multirow{2}{*}{Test-to-test}
& ASNet~\cite{cai2020messytable}
& \(0.525 \rightarrow \att{0.777}\)
& \(0.214 \rightarrow \att{0.106}\)
& \(0.672 \rightarrow \att{0.668}\)
& \(0.410 \rightarrow \att{0.388}\)
& \(0.163 \rightarrow \att{0.142}\) \\
& ASNet + ESC~\cite{cai2020messytable}
& \(0.576 \rightarrow \att{0.852}\)
& \(0.163 \rightarrow \att{0.059}\)
& \(0.724 \rightarrow \att{0.734}\)
& \(0.502 \rightarrow \att{0.501}\)
& \(0.227 \rightarrow \att{0.205}\) \\
\midrule
\multirow{2}{*}{Val-to-test}
& ASNet~\cite{cai2020messytable}
& \(0.525 \rightarrow \att{0.794}\)
& \(0.214 \rightarrow \att{0.096}\)
& \(0.672 \rightarrow \att{0.676}\)
& \(0.410 \rightarrow \att{0.402}\)
& \(0.163 \rightarrow \att{0.155}\) \\
& ASNet + ESC~\cite{cai2020messytable}
& \(0.576 \rightarrow \att{0.860}\)
& \(0.163 \rightarrow \att{0.055}\)
& \(0.724 \rightarrow \att{0.738}\)
& \(0.502 \rightarrow \att{0.505}\)
& \(0.227 \rightarrow \att{0.210}\) \\
\bottomrule
\end{tabular}
}
\caption{
Additional results on MessyTable object association.
Each cell reports the original value followed by the transformed value
(\(\mathrm{raw}\rightarrow\mathrm{transformed}\)).
}
\label{tab:supp_messytable}
\end{table*}

The results show that the same phenomenon appears beyond person association. On MessyTable, Sinkhorn-based post-processing substantially improves pairwise ranking metrics for both ASNet and ASNet+ESC. Under the val-to-test protocol, ASNet improves from \(0.525\) to \(0.794\) AP and from \(0.214\) to \(0.096\) FPR-95, while ACC changes only from \(0.672\) to \(0.676\). Similarly, ASNet+ESC improves from \(0.576\) to \(0.860\) AP and from \(0.163\) to \(0.055\) FPR-95, while ACC changes from \(0.724\) to \(0.738\). IPAA metrics also remain comparatively stable and do not show improvements comparable to the pairwise metrics. These results suggest that the mismatch between pairwise ranking metrics and assignment-level evaluation is not specific to multi-view person association, but also arises in general object association.

\section{Hungarian Matching Stability}
\label{app:hungarian_stability}

To more directly analyze whether the Sinkhorn-based post-processing changes the final
association decisions, we measure how often the Hungarian matching remains identical before and after post-processing. This diagnostic is complementary to ACC and IPAA: instead of only comparing aggregate assignment metrics, it directly checks whether the one-to-one matching induced by the affinity matrix changes at all.

\begin{table*}[t]
\centering
\small
\setlength{\tabcolsep}{4.0pt}
\renewcommand{\arraystretch}{0.95}
\resizebox{\textwidth}{!}{
\begin{tabular}{llccc}
\toprule
Dataset & Method
& AP \(\uparrow\)
& \(\Delta\)AP
& Unchanged Hungarian \(\uparrow\) \\
\midrule
\multirow{3}{*}{WILDTRACK}
& ReText~\cite{mamedov2026retext}
& \(0.528 \rightarrow \att{0.692}\)
& \(+0.164\) {\scriptsize(\(+31.1\%\))}
& \(470/840\) {\scriptsize(\(56.0\%\))} \\
& Self-MVA~\cite{chen2025learning}
& \(0.567 \rightarrow \att{0.953}\)
& \(+0.386\) {\scriptsize(\(+68.1\%\))}
& \(617/840\) {\scriptsize(\(73.5\%\))} \\
& OSNet + ESC~\cite{cai2020messytable}
& \(0.595 \rightarrow \att{0.809}\)
& \(+0.214\) {\scriptsize(\(+36.0\%\))}
& \(449/840\) {\scriptsize(\(53.5\%\))} \\
\midrule
\multirow{3}{*}{MVOR}
& ReText~\cite{mamedov2026retext}
& \(0.718 \rightarrow \att{0.894}\)
& \(+0.176\) {\scriptsize(\(+24.5\%\))}
& \(207/213\) {\scriptsize(\(97.2\%\))} \\
& OSNet + ESC~\cite{cai2020messytable}
& \(0.763 \rightarrow \att{0.884}\)
& \(+0.121\) {\scriptsize(\(+15.9\%\))}
& \(197/213\) {\scriptsize(\(92.5\%\))} \\
& Self-MVA~\cite{chen2025learning}
& \(0.840 \rightarrow \att{0.940}\)
& \(+0.100\) {\scriptsize(\(+11.9\%\))}
& \(207/213\) {\scriptsize(\(97.2\%\))} \\
\bottomrule
\end{tabular}
}
\caption{
Hungarian assignment stability under Sinkhorn-based post-processing in the val-to-test protocol. Each AP cell reports the original value followed by the transformed value (\(\mathrm{raw}\rightarrow\mathrm{transformed}\)). The last column reports the number and percentage of affinity matrices whose Hungarian matching is identical before and after post-processing.
}
\label{tab:supp_hungarian_stability}
\end{table*}

The results provide a direct view of the mechanism behind the metric mismatch. On MVOR, AP improves substantially while the Hungarian matching remains unchanged for more than \(90\%\) of image-pair affinity matrices for all evaluated methods. For example, ReText improves from \(0.718\) to \(0.894\) AP while preserving the matching in \(97.2\%\) of matrices, and Self-MVA improves from \(0.840\) to \(0.940\) AP with the same \(97.2\%\) unchanged rate. On WILDTRACK, the unchanged rate is lower, indicating that post-processing also alters some assignments. Nevertheless, AP gains remain much larger and more systematic than the corresponding assignment-level changes reported in the main paper. This supports our central claim: pairwise ranking metrics can be strongly affected by score reshaping even when the underlying Hungarian assignment is preserved for a large fraction of affinity matrices.